\pdfoutput=1

\documentclass[11pt]{article}

\usepackage[]{EMNLP2022}

\usepackage{times}
\usepackage{latexsym}

\usepackage[T1]{fontenc}

\usepackage[utf8]{inputenc}

\usepackage{microtype}

\usepackage{inconsolata}

\newcommand{\wanli}{\textsc{WaNLI}\xspace}

\title{\wanli: Worker and AI Collaboration for \\ 
Natural Language Inference Dataset Creation}

\newcommand{\aspace}{\hspace{1em}}
\newcommand{\uw}{$^{\heartsuit}$}
\newcommand{\usc}{$^{\diamondsuit}$}
\newcommand{\aiTwo}{$^{\clubsuit}$}
\author{
    Alisa Liu\uw\aspace 
    Swabha Swayamdipta\aiTwo\usc\aspace
    Noah A. Smith\uw\aiTwo \aspace 
    Yejin Choi\uw\aiTwo \aspace \\
    \uw Paul G.\ Allen School of Computer Science \& Engineering, University of Washington \\
    \aiTwo Allen Institute for Artificial Intelligence \quad \usc University of Southern California\\
    \texttt{alisaliu@cs.washington.edu}
}

\begin{document}
\maketitle
\begin{abstract}
A recurring challenge of crowdsourcing NLP datasets at scale is that human writers often rely on repetitive patterns when crafting examples, leading to a lack of linguistic diversity.
We introduce a novel approach for dataset creation based on \textbf{worker and AI collaboration}, which brings together the generative strength of language models and the evaluative strength of humans.
Starting with an existing dataset, MultiNLI for natural language inference (NLI), our approach uses dataset cartography to automatically identify examples that demonstrate challenging reasoning patterns, and instructs GPT-3 to compose new examples with similar patterns.
Machine generated examples are then automatically filtered, and finally revised and labeled by human crowdworkers.
The resulting dataset, \wanli, consists of 107,885 NLI examples and presents unique empirical strengths over existing NLI datasets. 
Remarkably, training a model on \wanli improves performance on eight out-of-domain test sets we consider, including by 11\% on HANS and 9\% on Adversarial NLI, compared to training on the $4\times$ larger MultiNLI.
Moreover, it continues to be more effective than MultiNLI augmented with other NLI datasets.
Our results demonstrate the promise of leveraging natural language generation techniques and re-imagining the role of humans in the dataset creation process.
\end{abstract}


\begin{figure}[t]
    \centering
    \includegraphics[width=0.9\columnwidth]{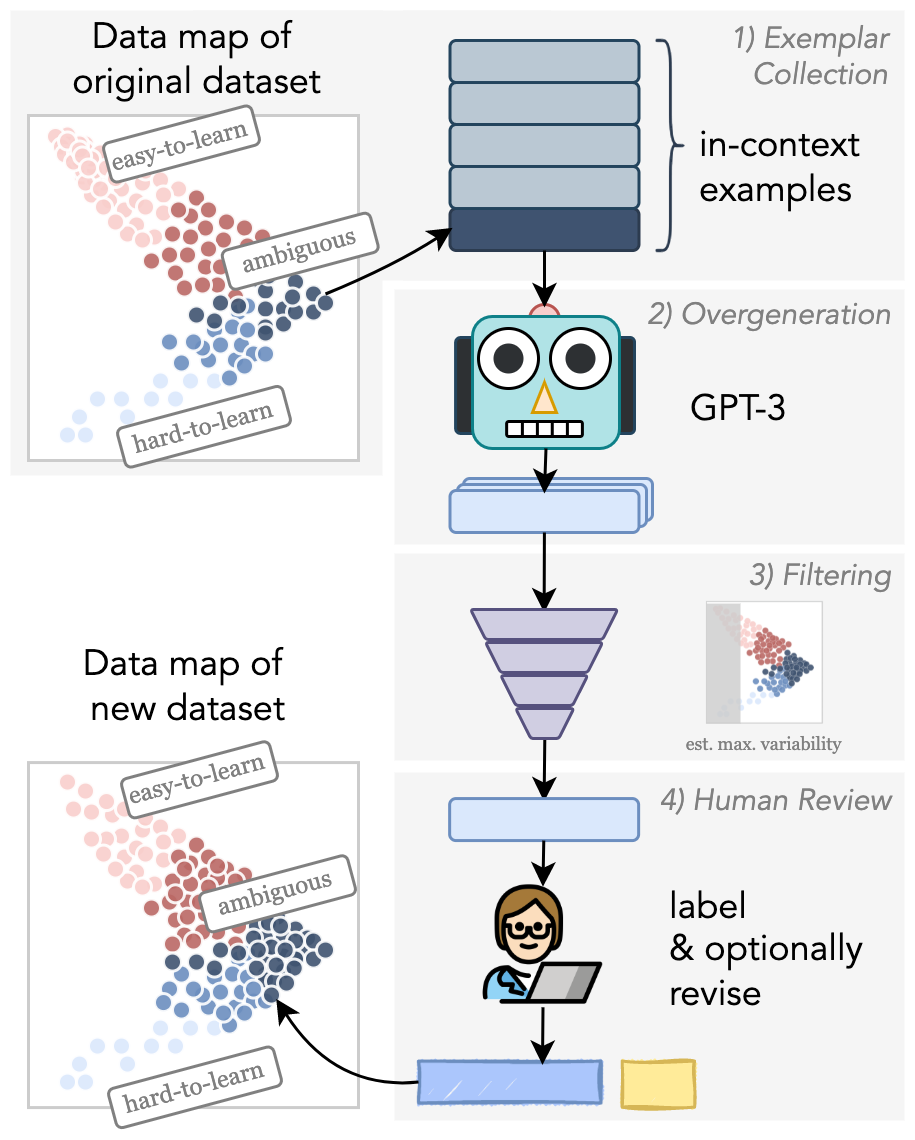}
    \caption{
    An illustration of our pipeline for creating \wanli.
    Starting with a data map \cite{swayamdipta-etal-2020-dataset} of an existing dataset relative to a trained model, (1) we automatically identify pockets of data instances exemplifying challenging reasoning patterns.
    Next, (2) we use GPT-3 to generate new instances with the same pattern. 
    These generated examples are then (3) automatically filtered via a metric we introduce inspired by data maps, and (4) given to human annotators to assign a gold label and optionally revise.}
    \label{fig:figure_1}
\end{figure}

\section{Introduction}
\label{sec:intro}
As much as large-scale crowdsourced datasets have expedited progress on various NLP problems, a growing body of research has revealed fundamental limitations in existing datasets: they are often flooded with repetitive and spurious patterns, rather than covering the broad range of linguistic phenomena required by the task \cite{bowman-dahl-2021-will}.
This leads to models that seem to achieve human-level performance on in-domain test sets, yet are brittle when given out-of-domain or adversarial examples \cite{ribeiro-etal-2020-beyond, glockner-etal-2018-breaking}.

We attribute this problem to an inherent challenge in the crowdsourcing design---the prevalent paradigm for creating large-scale NLP datasets---where a relatively small number of workers create a massive number of free text examples. 
While human annotators are 
generally reliable for writing \emph{correct} examples, crafting \emph{diverse and creative} examples at scale can be challenging. Thus, crowdworkers often resort to a limited set of writing strategies for speed, at the expense of diversity \cite{geva-etal-2019-modeling,gururangan-etal-2018-annotation}.
When models overfit to such repetitive patterns, they fail to generalize to out-of-domain examples where these patterns no longer hold \cite{geirhos-etal-2020-shortcut}. 

On the other hand, there has been remarkable progress in open-ended text generation based on massive language models \cite[i.a.]{brown-etal-2020-language,raffel-etal-2020-exploring}.
Despite known deficiencies such as incoherence or repetition \cite{dou-etal-2021-scarecrow}, these models often produce human-like text \cite{clark-etal-2021-thats} and show potential for creative writing tasks \cite{lee-etal-2022-coauthor}.
Importantly, these models are capable of replicating a pattern given just a few examples in context (\citealp[GPT-3]{brown-etal-2020-language}).

In this paper, we introduce a novel approach for dataset creation which brings together the generative strength of language models and the evaluative strength of humans through \textbf{human and machine collaboration} (\S\ref{sec:method}).
The key insight of our approach is that language models can create new examples by replicating linguistic patterns that are valuable for training, without necessarily ``understanding'' the task itself.
Illustrated in \autoref{fig:figure_1}, our pipeline starts with an existing dataset.
We use dataset cartography from \citet{swayamdipta-etal-2020-dataset} to automatically identify pockets of examples that demonstrate challenging reasoning patterns relative to a trained model.
Using each group as a set of in-context examples, we leverage a pretrained language model to generate new examples likely to have the same pattern (see \autoref{tab:examples}).
We then propose a novel metric, building on dataset cartography, to automatically filter generations that are most likely to aid model learning.
Finally, we validate the generated examples by subjecting them to human review, where crowdworkers assign a gold label and (optionally) revise for quality.

\begin{table*}[t]
    \centering
    \resizebox{\textwidth}{!}{%
    \begin{tabular}{lll}
    \toprule
        \textbf{Seed MultiNLI example} & \textbf{Generated \wanli Example} & \textbf{Label \& Reasoning}  \\\midrule
        \makecell[l]{
        P: \textit{\textbf{5 percent}} probability that each part will be defect free.\\
        H: Each part has a \textit{\textbf{95 percent}} chance of
        having a defect.
        } & \makecell[l]{
        P: \textit{\textbf{1 percent}} of the seats were vacant.\\
        H: \textit{\textbf{99 percent}} of the seats were occupied.}
        &  \makecell[l]{\textit{\colorbox{ent}{Entailment}}\\ Set complements}\\ \midrule
        \makecell[l]{
        P: The artisans, tradespeople, and providers of entertainment \\
        (reputable and not so reputable) lived downtown on the\\ reclaimed marshlands north and east, in the area still known\\ 
        as Shitamachi.\\
        H: \textit{\textbf{The only place}} where artisans, tradespeople and entertainers\\
        could live was in the marshlands to the north and east.
        }
        & \makecell[l]{
        P: At the time of the Revolution, the old port of Marseille\\
        was a great center of shipbuilding and commerce.\\
        H: \textit{\textbf{The only place}} where ships were built was in the old\\
        port of Marseille.
        }& \makecell[l]{\textit{\colorbox{neu}{Neutral}}\\Hypothesis introduces\\exclusivity}\\\midrule
        \makecell[l]{
        P: To the \textit{\textbf{south}}, in the Sea of Marmara, lie the woods and \\
        beaches of the Princes' Islands.\\
        H: In the \textit{\textbf{north}} is the Sea of Marmara where there are \\mountains to climb.}
        & \makecell[l]{
        P: From the park's \textit{\textbf{southern entrance}}, follow the avenue\\
        \textit{\textbf{south}} to the Hotel de Ville.\\
        H: From the park's \textit{\textbf{northern entrance}}, follow the avenue\\
        \textit{\textbf{north}} to the Hotel de Ville.}
        & \makecell[l]{\textit{\colorbox{con}{Contradiction}}\\Reversing cardinal\\ directions}\\\midrule
        \makecell[l]{
        P: Democrats released documents indicating that
        Republicans \\
        sold big political donors meals with the party's leaders in \\ 
        federal buildings in 1995.\\
        H: \textit{\textbf{It is illegal}} for a party to solicit products to donors. \\
        }
        &\makecell[l]{
        P: In the late 1960s, students at a university in Wisconsin\\
        tried to organize a union.\\
        H: \textit{\textbf{It was illegal}} for the students to organize a union.
        }
        & \makecell[l]{
        \textit{\colorbox{neu}{Neutral}}\\
        Illegal things\\
        can happen}\\\midrule
        \makecell[l]{
        P: She ducked \textit{\textbf{and}} parried the blow.\\
        H: She ducked \textit{\textbf{to}} miss the blow.
        } 
        & \makecell[l]{
        P: She stepped on the brake \textit{\textbf{and}} the car came to a stop.\\
        H: She stepped on the brake \textit{\textbf{to}} stop the car.
        }
        & \makecell[l]{\textit{\colorbox{ent}{Entailment}}\\Implied intention} \\\midrule
        \makecell[l]{
        P: To build a worldclass finance organization and help\\
        achieve better business outcomes, each of the organizations \\
        we examined \textbf{\textit{set an agenda for transforming}} the finance \\
        organization by defining a  shared vision -i.e.\\
        H: \textit{\textbf{The transformation was a disaster}} and the entire \\
        organization had to be scrapped.}
        & \makecell[l]{
        P: In order to help improve customer service, \textit{\textbf{I suggested}} \\
        \textit{\textbf{that they send a representative}} to our office to discuss \\
        our concerns.\\
        H: \textit{\textbf{The representative}} sent to our office \textbf{\textit{did not solve}} \\
        \textbf{\textit{our problems}} and we lost a lot of business.} 
        &\makecell[l]{
        \textit{\colorbox{neu}{Neutral}} \\
        Intended goals may\\
        not actualize} \\\midrule
        \makecell[l]{
        P: Salinger \textit{\textbf{wrote}} similar letters \textbf{\textit{to}} other young female writers.\\
        H: Other young female writers \textit{\textbf{received}} similar letters \textbf{\textit{from}}\\
        Salinger as well.}
        &\makecell[l]{
        P: The three schools \textit{\textbf{have}} a number of students who are \\
        from families with no history of financial difficulties.\\
        H: Families with no history of financial difficulties \\
        \textbf{\textit{send}} their children to the three schools.} &
        \makecell[l]{
        \textit{\colorbox{ent}{Entailment}} \\
        Substituting a verb \\
        with a different \\
        subcategorization frame} \\
    \bottomrule
    \end{tabular}}
    \caption{Seed MultiNLI examples, and corresponding \wanli examples generated by GPT-3.
    P stands for premise, H for hypothesis. 
    The seed example is ``ambiguous'' according to the definitions of \citet{swayamdipta-etal-2020-dataset}, discussed in \S\ref{sec:method}. 
    The remaining in-context examples (shown in Appendix \ref{subsec:context_examples}) share the same pattern and are found using distance in \texttt{[CLS]} embeddings of a trained task model. 
    The reasoning is a short description of the pattern we observe from the group, and which is successfully repeated in the generated example.}
    \label{tab:examples}
\end{table*}


We demonstrate the effectiveness of our approach on the task of natural language inference (NLI), which determines whether a premise entails (i.e., implies the truth of) a hypothesis, both expressed in natural language. 
Despite being one of the most resource-available tasks in NLP, analysis and challenge sets repeatedly demonstrate the limitations of existing datasets and the brittleness of NLI models trained on them \cite{gururangan-etal-2018-annotation, poliak-etal-2018-hypothesis, tsuchiya-2018-performance}.
Using MultiNLI \cite{williams-etal-2018-broad} as our original dataset, we use our pipeline to create a dataset of 107,885 examples, which we call \textbf{W}orker-and-\textbf{A}I \textbf{NLI} (\wanli).\footnote{Pronounced wan-li like the Chinese characters \chinese{万}\chinese{理}, as in \textit{ten thousand reasoning}. A demo, data, and code are available at \url{https://wanli.allenai.org/}.}

Remarkably, empirical results demonstrate that \textit{replacing} MultiNLI supervision with \wanli (which is $4$ times smaller) improves performance on eight different out-of-domain test sets, including datasets that are converted to the NLI format from downstream tasks such as question-answering and fact verification (\S\ref{sec:extrinsic-eval}).
This result holds even when augmenting MultiNLI with other NLI datasets and recently proposed augmentation sets. 
Moreover, including \wanli in the training data can help improve performance on certain in-domain test sets. 
We then analyze \wanli and show that it has fewer previously documented spurious correlations than MultiNLI (\S\ref{sec:analysis}), and provide insights into the collaborative framework (\S\ref{sec:discussion}).

Our approach contrasts with previous instruction-based generation of dataset examples \cite{schick-schutze-2021-generating, west-etal-2021-symbolic}, which require the model to understand the task from context, fundamentally limiting the complexity of generated output to what is accessible by the model. 
Moreover, our human-in-the-loop approach is \textit{collaborative}, rather than \textit{adversarial} \cite{dinan-etal-2019-build,nie-etal-2020-adversarial,bartolo-etal-2020-beat}.
Overall, we leverage the best of both worlds: a powerful model's ability to efficiently generate diverse examples, and humans' ability to improve and ensure the quality of generations. 

Our worker-AI collaborative approach is more scalable compared to the traditional crowdsourcing framework.
Our approach is generalizable, allowing for rejuvenating datasets on many different classification tasks, especially when performance seems to stagnate due to overfitting to popular benchmarks \cite{recht2019imagenet}.
Our work shows the promise of leveraging language models in a controlled way to aid the dataset creation process, and we encourage the community to think of dataset curation as an AI challenge itself.
\section{Worker-AI Collaborative Dataset Creation for NLI}
\label{sec:method}

We describe our four-stage approach for dataset creation based on worker and AI collaboration.
In this work, we apply it to the task of natural language inference (NLI), which involves predicting whether a premise \textit{entails}, \textit{contradicts} or is \textit{neutral} to a hypothesis.
NLI has broad applicability in NLP: it has proven useful for pretraining \cite{clark-etal-2019-boolq,phang-etal-2018-sentence}, and can be applied to verify candidate answers in question-answering \cite{chen-etal-2021-nli-models} or factuality of generated summaries \cite{maynez-etal-2020-faithfulness}.

Our approach requires as prerequisites an initial dataset $\mathcal D_0$ and a strong task model $\mathcal M$ trained on $\mathcal D_0$. 
We use MultiNLI \cite{williams-etal-2018-broad}, a large-scale multi-genre NLI dataset, as $\mathcal D_0$.
We finetune RoBERTa-large \cite{liu-etal-2019-roberta} on MultiNLI for our task model $\mathcal M$ (training details in \autoref{sec:modeling_details}).

As an overview, we first automatically \textbf{collect} groups of examples exemplifying challenging reasoning patterns in $\mathcal D_0$ relative to $\mathcal M$, using data maps (\citealp{swayamdipta-etal-2020-dataset}; Stage 1, see \S\ref{subsec:exemplar_collection}).
Then we \textbf{overgenerate} similar examples by leveraging the pattern replication capabilities of GPT-3 \cite{brown-etal-2020-language} (Stage 2; \S\ref{subsec:overgeneration}). 
While GPT-3 can generate examples efficiently, it may not reliably replicate the desired pattern and its output quality will not be uniform.
We address this by automatically \textbf{filtering} the generated examples using a metric derived from data maps (Stage 3; \S\ref{subsec:filtering}).
We finally subject the collected data to \textbf{human review}, in which crowdworkers optionally revise examples and assign gold labels (Stage 4; \S\ref{sec:human_review}).


\paragraph{Dataset Cartography.} 
A key component of our pipeline is inspired by data maps \cite{swayamdipta-etal-2020-dataset}, which automatically reveal different regions in a dataset, w.r.t.\ the behavior of a classification model during training. 
These include \textit{easy-to-learn} examples which the model consistently predicts correctly through training, \textit{hard-to-learn} examples on which it is consistently incorrect, and \textit{ambiguous} examples for which the model's confidence in the correct answer exhibits high \textit{variability} across train epochs.
Our pipeline focuses on \textit{ambiguous} examples, which were shown to lead to more robust models. 
Additionally, ambiguous examples contain fewer spurious correlations \cite{gardner-etal-2021-competency}, suggesting that they capture under-represented counterexamples to spurious correlations.
Indeed, such counterexamples take more epochs of training to learn and are crucial for generalization \cite{tu-etal-2020-empirical}, providing a potential explanation for why they appear ambiguous across early epochs and lead to more robust models. 

\subsection{Stage 1: Collection of Exemplars}
\label{subsec:exemplar_collection}

In this stage, we automatically collect groups of examples from $\mathcal D_0$ which represent linguistic patterns we wish to include in the target dataset.
We begin with a seed example $(x_i,y_i)\in\mathcal D_0$ belonging to the most ambiguous $p=25\%$ relative to $\mathcal M$.\footnote{For exemplar collection, we exclude the telephone genre of MultiNLI, which consists of telephone conversation transcripts, due to their low fluency and ill-defined entailment relationships. During pilots, we found that generated examples mimicking telephone conversations would require crowdworkers to revise low-quality text for basic fluency.}


To generate a new example with the same reasoning pattern, we wish to leverage the ability of GPT-3 \cite{brown-etal-2020-language} for in-context learning; hence, we need to first collect examples that test a similar kind of reasoning to $x_i$.
To do this, we use the \texttt{[CLS]} token representation of each example relative to the \textit{task} model $\mathcal M$, and find the $k=4$ nearest neighbors via cosine similarity to $x_i$ that \textit{have the same label}.
Detailed qualitative inspection shows that the nearest neighbors in this representation space tend to capture a human-interpretable similarity in the \textit{reasoning} required to solve an example, rather than lexical or semantic similarity (examples in \autoref{tab:examples}).

\citet{han-tsvetkov-2021-influence-tuning} give another interpretation for this approach: for examples with the same label, the similarity of \texttt{[CLS]} token embeddings actually represents the similarity of \textit{gradient updates} in the row of the final projection layer corresponding to that label.
Thus, two examples are close if training on them would ``update'' the final layer of the model similarly.

By automatically identifying areas for augmentation, our method does not require any prior knowledge of challenging patterns and makes our method tractable for building on top of large-scale datasets.
Nonetheless, exemplar collection could potentially be approached in different ways (e.g., through expert curation or category labels).

\subsection{Stage 2: Overgeneration}
\label{subsec:overgeneration}

\begin{figure}
    \centering
    \includegraphics[width=0.9\columnwidth]{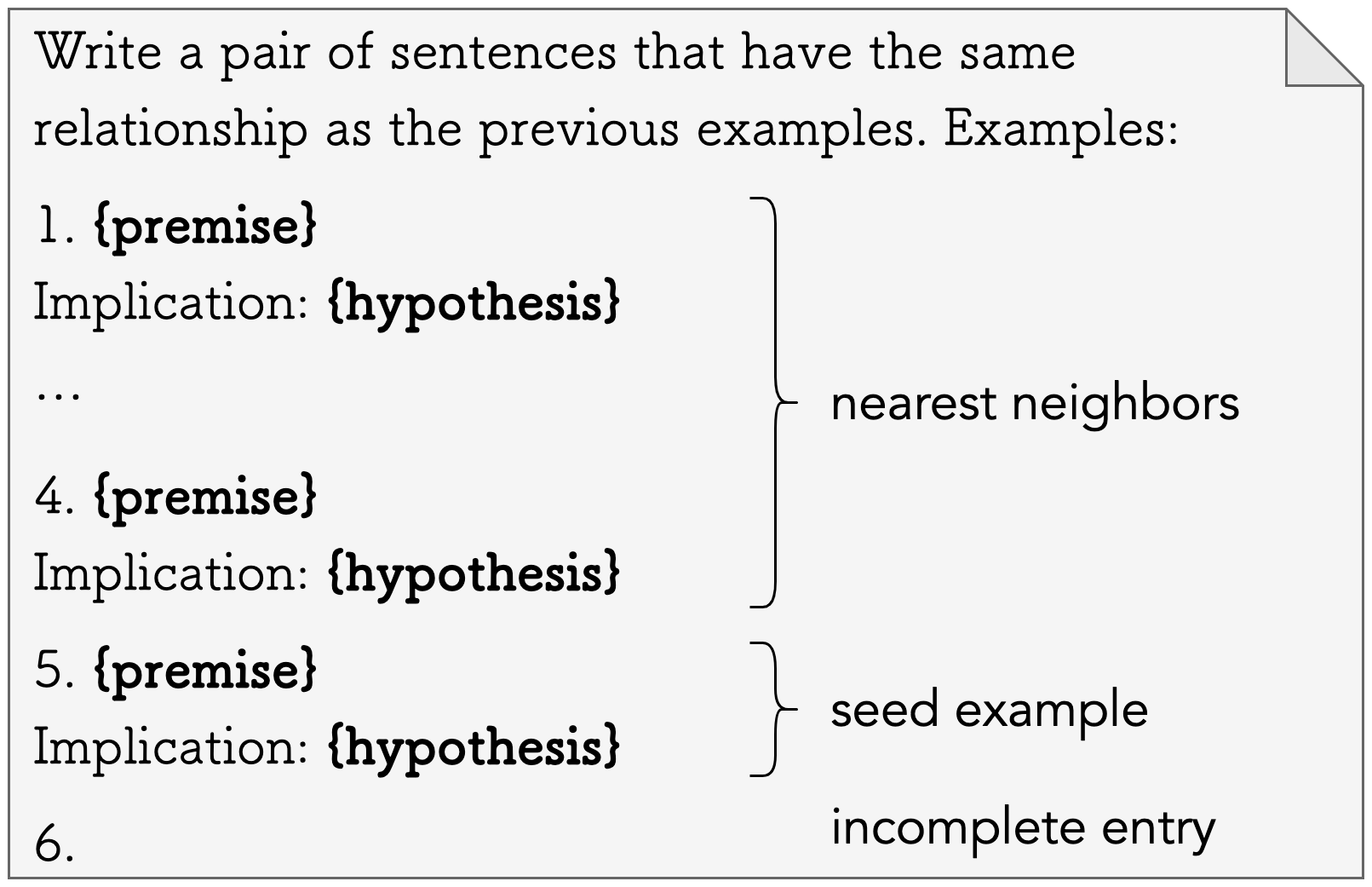}
    \caption{
    Prompt template instructing GPT-3 to generate a new example, given a set of in-context examples. 
    To separate the premise and hypothesis, the word ``Implication'' is used for entailment examples (shown here), ``Possibility'' for neutral examples, and ``Contradiction'' for contradiction examples.}
    \label{fig:template}
\end{figure}

Given an automatically extracted group of $k+1$ examples from the original dataset $\mathcal D_0$, we construct a natural language context (prompt) for a left-to-right language model; in this work, we use GPT-3 Curie (the second-largest GPT-3 model).
The prompt template we use is shown in \autoref{fig:template}, where we order the examples in \textit{increasing} similarity to the seed example.

Note that our method leverages GPT-3 in way that is distinct from its typical usage in few-shot settings, where given examples demonstrating a task, GPT-3 performs the task on a new, unlabeled example.
Here, we instead give GPT-3 examples representing a particular \textit{slice} of the task, and ask GPT-3 to \textit{generate} a new example in the same slice. 

For each context, we sample from GPT-3 to create $n=5$ distinct examples. 
We use top-$p$ decoding \cite{holtzman-etal-2020-curious} with $p=0.5$
(additional details in Appendix \ref{subsec:generation_hyperparams}).
Although generated examples at this stage could be assumed to share label of its $k+1$ in-context examples, we instead consider the resulting dataset $\mathcal{D}\textsubscript{gen}=\{x_i\}_i$ at the end of Stage 1 to be \emph{unlabeled}.

\subsection{Stage 3: Automatic Filtering}
\label{subsec:filtering}

In this step, we wish to filter generated examples from Stage 2 to retain those that are the most ambiguous with respect to $\mathcal M$. 
However, computing ambiguity for an example requires that it be a part of the original training set, whereas we wish to estimate the ambiguity of an \textit{unlabeled} example \textit{without} additional training.
Thus we introduce a new metric called \textbf{estimated max variability}, which measures the worst-case spread of predictions on an example $x_i$ across checkpoints of a trained model. 
Let $E$ be the total epochs in training, $\mathcal Y$ the label set, and $p_{\theta^{(e)}}$ the probability assigned with parameters $\theta^{e}$ at the end of the $e$-th epoch. 
We define the estimated max variability as:
\begin{equation}
    \sigma_i=\max_{y\in \mathcal Y}\sigma\left(\left\{p_{\theta^{(e)}}(y\mid x_i)\right\}_{e\in E}\right),
\end{equation}
\noindent where $\sigma$ is the standard deviation function. 

Concretely, we \textit{retroactively} compute the prediction from each saved epoch of $\mathcal M$ on $x_i$.  
The only assumption made is that the single example, if it had been a part of the training set, would have made a negligible difference on each model checkpoint (at least as observed through its posterior probabilities).\footnote{Indeed, we find a high correlation between variability and estimated max variability; see \autoref{sec:max_variability}.}
In taking a maximum across labels, we consider $x_i$ to be ambiguous as long as $\mathcal M$ is undecided on \textit{any} label $\in \mathcal{Y}$.

We first employ simple heuristics to discard examples exhibiting observable failure cases of GPT-3.
Specifically, we discard examples where 1) the premise and hypothesis are identical, modulo punctuation or casing, 2) the generated example is an exact copy of an in-context example, 3) the example contains some phrases from the instruction (e.g., ``pair of sentences''), or 4) the premise or hypothesis is shorter than 5 characters. 
Then, we compute the estimated max variability for the remaining examples with respect to $\mathcal M$, and retain an equal number of examples from each (intended) label class with the highest max variability, to create a dataset $\mathcal D\textsubscript{filtered}$ that is half the size of $\mathcal D\textsubscript{gen}$.


\subsection{Stage 4: Human Review}
\label{sec:human_review}

As the final stage of our pipeline, we recruit human annotators on Amazon Mechanical Turk to review each unlabeled example $x_i\in\mathcal D\textsubscript{filtered}$. (Details about crowdworkers and guidelines in \autoref{sec:human_annotation}.)
The annotator may optionally revise $x_i$ to create a higher-quality example $x_i'$, or let $x_i' = x_i$. 
Either way, they assign a label $y_i$.
When revising examples, we asked annotators to preserve the intended meaning as much as possible through minimal revisions.\footnote{In pilots, we found that when annotators exercised too much freedom in revision, they often re-introduced the same artifacts that have been well-documented in NLI.}
However, if an example would require a great deal of revision to fix \textit{or} if it could be perceived as offensive, they should discard it.
This results in the labeled dataset $\mathcal D\textsubscript{collab}=\{(x_i',y_i)\}_i$.


Crowdworkers annotate a total of 118,724 examples, with two distinct workers reviewing each example.
For examples that both annotators labeled without revision, we achieved a Cohen's $\kappa$ of $0.60$, indicating substantial agreement.
To create the final dataset, we discard an example if \textit{either} annotator chose to discard it, and we keep a revision only if \textit{both} annotators revise an example (and choose a revision uniformly at random).
When both annotators label the example as-is but choose different labels, we sample one of the two labels uniformly at random. 
The rationale for this is discussed in Appendix \ref{subsec:disagreement_more}.
This leads to a labeled dataset of 107,885 examples (90.87\% of all annotated examples, with the remaining discarded).
Of the labeled examples, 3.54\% were revised. 

We randomly split the data into a train and test sets.
Key dataset statistics are summarized in \autoref{tab:dataset_statistics}.
Unlike MultiNLI, \wanli is not label-balanced; see \S\ref{subsec:in-context} for a discussion.

In general, we believe the role 
of revision depends on the quality of machine-generated examples.
Indeed, we need to strike a balance between leveraging human capabilities and avoiding the re-emergence of annotation artifacts that may come with too much freedom in revision.

\begin{table}[t]
    \centering
    \small
    \resizebox{\columnwidth}{!}{%
    \begin{tabular}{lrc}
    \toprule
        \textbf{Split} & \textbf{Size} & \textbf{Label distribution} (E/N/C) \\\midrule
        Train & 102,885 & 38,511 / 48,977 / 15,397 \\
        Test & 5,000 & 1,858 / 2,397 / 745 \\
    \bottomrule
    \end{tabular}}
    \caption{\wanli dataset statistics.}
    \label{tab:dataset_statistics}
\end{table}


\begin{table*}[t]
    \centering
    \resizebox{\textwidth}{!}{%
    \begin{tabular}{clcqstuvwxyz}
        \toprule
        &&&\multicolumn{9}{c}{\textbf{Test Set}}\\\cmidrule{4-12}
        &&& \multicolumn{1}{c}{\textbf{Diagnostics}} & \multicolumn{1}{c}{\textbf{HANS}*} & \multicolumn{1}{c}{\textbf{QNLI}*} & \multicolumn{1}{c}{\textbf{WNLI}*} & \multicolumn{1}{c}{\textbf{NQ-NLI}*} & \multicolumn{1}{c}{\textbf{ANLI}} & \multicolumn{1}{c}{\textbf{FEVER-NLI}} & \multicolumn{1}{c}{\textbf{BIG-Bench}*} & \multicolumn{1}{c}{\textbf{\wanli}} \\
        && \multicolumn{1}{r}{Data size} & 1104 & 30K & 5266 & 706 & 4855 & 3200 & 20K & 3324 & 5000 \\\midrule
        \multirow{12}{*}{\rotatebox[origin=c]{90}{\textbf{Training Set}}} & MNLI & 393K & 68.47 & 78.08 & 52.69 & 56.09 & 62.34 & 32.37 & 68.29 & 64.68 & 64.62\\\cmidrule{2-12}
        &  MNLI + Tailor & 485K & 67.75 & 79.03 & 54.89 & 56.23 & 63.83 & 32.87 & 68.75 & 72.38 & 64.27 \\
        & MNLI + Z-Aug & 754K & 66.39 & 80.52 & 57.72 & 55.52 & 62.30 & 33.37 & 68.73 & 66.12 & 64.78 \\
        &  MNLI $\diamond$ ANLI & 393K & 67.75 & 79.90 & 68.74 & 60.48 & 62.49 & \cellcolor[gray]{0.8}{54.59} & \cellcolor[gray]{0.8}{72.30} & 72.32 & \textbf{65.96} \\
        &  MNLI + ANLI & 556K & 66.84 & 77.94 & 62.41 & 57.08 & 62.84 & \cellcolor[gray]{0.8}{53.84} & \cellcolor[gray]{0.8}{72.30} & 71.11 & 65.93 \\
        & MNLI $\diamond$ FEVER-NLI & 393K & 66.75 & 76.50 & 56.70 & 57.08 & 61.81 & 35.65 & \cellcolor[gray]{0.8}{76.83} & 58.39 & 63.31 \\
        & MNLI + FEVER-NLI & 601K & 67.57 & 76.05 & 52.90 & 54.95 & 63.02 & 35.37 & \cellcolor[gray]{0.8}{76.93} & 64.65 & 64.53 \\
        & MNLI + SNLI + ANLI & 943K & 68.75 & 78.65 & 63.38 & 58.49 & 62.94 & \cellcolor[gray]{0.8}{54.21} & \cellcolor[gray]{0.8}{72.02} & 71.05 & 65.10 \\\cmidrule{2-12}
        & MNLI $\diamond$ \textbf{\wanli} & 393K & 71.01 & 83.10 & 77.00 & 61.89 & 62.94 & 36.46 & \textbf{71.14} & 76.17 & \cellcolor[gray]{0.8}{75.49} \\
        & MNLI + \textbf{\wanli} & 496K & 71.64 & 82.00 & 68.40 & 60.05 & 63.21 & 36.78 & 70.79 & 70.81 & \cellcolor[gray]{0.8}{75.26}\\
        & \textbf{\wanli} & 103K & \textbf{72.73} & \textbf{89.28} & \textbf{81.40} & \textbf{67.28} & \textbf{64.18} & \textbf{41.12} & 70.13 & \textbf{85.19} & \cellcolor[gray]{0.8}{75.40}\\
        \bottomrule
    \end{tabular}}
    \caption{
    Empirical comparison of different training sets for RoBERTa-large, for generalization to out-of-distribution (OOD) challenge sets. 
    Gray cells mark settings that do not represent an OOD challenge.
    \textbf{Top}: Training on MultiNLI alone.
    \textbf{Middle}: Comparison of combination schemes with MultiNLI. 
    We consider two data combination strategies, augmentation (+), and random replacement ($\diamond$), where the resulting dataset size is unchanged.
    \textbf{Bottom}: Training sets that include \wanli.
    The highest accuracy on each test set (excluding gray cells) is bolded.
    Test sets with * contain two label classes: entailment and non-entailment.}
    \label{tab:results}
\end{table*}

\section{Training NLI Models with \wanli}
\label{sec:extrinsic-eval}

We finetune different copies of RoBERTa-large \cite{liu-etal-2019-roberta} on different training sets, and evaluate each resulting model's performance on a large suite of NLI challenge sets.
Given that the challenge sets were constructed independently of MultiNLI or \wanli, we consider them out-of-distribution (OOD) for both training datasets.

\subsection{NLI Test Suite}
\label{sec:nli-test-suite}

The NLI challenge sets come from a wide array of domains, methodologies (e.g., crowdsourcing, expert curation, generation), and initial task formats (e.g., question-answering, fact verification).\footnote{We evaluate on the development set for every dataset, except for Winograd NLI, where we combine the train and development set for greater statistical power, and Adversarial NLI, where we use the test set as the labels were not hidden.}

\vspace{.1cm}\noindent\textbf{NLI Diagnostics} \cite{wang-etal-2018-glue} is a manually-curated test set that evaluates a variety of linguistic phenomena using naturally-occurring sentences from several domains.

\vspace{.1cm}\noindent\textbf{HANS} \cite{mccoy-etal-2019-right} targets unreliable syntactic heuristics based on lexical overlap between the premise and hypothesis.

\vspace{.1cm}\noindent\textbf{QNLI} was adapted from the Stanford Question-Answering Dataset \cite{rajpurkar-etal-2016-squad} by the GLUE benchmark \cite{wang-etal-2018-glue}. Each example consists of a premise that is a sentence, and a hypothesis that is a question, which is entailed if the question is answered by the premise.

\vspace{.1cm}\noindent{\textbf{Winograd NLI}} was adapted by the GLUE benchmark from the Winograd Schema Challenge \cite{levesque-etal-2011-winograd}, which tests correct coreference via common sense. 
To convert this dataset to NLI, an entailed hypothesis is formed by substituting a correct referent and a non-entailed hypothesis is formed by substituting an incorrect referent.

\vspace{.1cm}\noindent\textbf{Adversarial NLI}  \cite[ANLI;][]{nie-etal-2020-adversarial} is an adversarially-constructed dataset where crowdworkers are instructed to write examples that stump existing models.
Examples are collected in three rounds that progressively increase in difficulty, with model adversaries trained on MultiNLI, SNLI \cite{bowman-etal-2015-large}, FEVER-NLI (discussed below), as well as ANLI sets from earlier rounds.

\vspace{.1cm}\noindent\textbf{Natural Questions NLI} \cite[NQ-NLI,][]{chen-etal-2021-nli-models} is created from the Natural Questions QA dataset \cite{kwiatkowski-etal-2019-natural}. The premise is a \textit{decontextualized} sentence from the original context; the hypothesis consists of a question and answer candidate converted into declarative form.

\vspace{.1cm}\noindent\textbf{FEVER NLI} is adapted from the FEVER fact verification dataset \cite{thorne-etal-2018-fever}, and introduced along with ANLI.
In each example, the premise is a short context from Wikipedia, and the hypothesis is a claim that is either supported (entailed), refuted (contradicted), or neither (neutral).

\vspace{.1cm}\noindent\textbf{BIG-Bench NLI} is a combination of four datasets from BIG-Bench \cite{srivastava-etal-2022-bigbench} about entailment: Analytic Entailment, Epistemic Reasoning, Disambiguation QA, Presuppositions NLI.

\subsection{Training Datasets}
\label{sec:training_datasets}
In addition to stand-alone \wanli and MultiNLI, we also consider combining MultiNLI with other NLI datasets. We use the train sets of SNLI \cite{bowman-etal-2015-large}, ANLI, and FEVER-NLI, as well as the augmentation set generated via \textsc{Tailor} \cite{ross-etal-2022-tailor}, which perturbed SNLI hypotheses to create examples with high lexical overlap between the premise and hypothesis, and the augmentation set Z-Aug \cite{wu-etal-2022-generating}, which was created by generating in-distribution examples and filtering them based on spurious correlations.

We consider two schemes for combining datasets $\mathcal A$ and $\mathcal B$: 1) \textbf{augmentation} ($\mathcal A+\mathcal B$), in which the two datasets are concatenated, and 2) \textbf{random replacement} ($\mathcal A\diamond\mathcal B$), where $\lvert \mathcal B\rvert$ examples from $\mathcal A$ are randomly swapped out and replaced with all examples from $\mathcal B$. 


\subsection{Results}
\label{sec:results}

Results are shown in \autoref{tab:results}.
When comparing MultiNLI (MNLI) and \wanli alone, training a model on \wanli instead of MultiNLI leads to better performance on every test set we consider, including by $4\%$ on Diagnostics, $11\%$ on HANS, and $9\%$ on Adversarial NLI.
This is remarkable given \wanli is $4\times$ smaller than MultiNLI, and contains primarily machine-written examples.

A \wanli-trained model continues to outperform baselines that combine MultiNLI with other NLI datasets and augmentation sets, in every OOD setting. 
This includes when comparing to a model trained on $9\times$ more data from three existing NLI datasets, MNLI $+$ SNLI $+$ ANLI.
The consistent advantage of \wanli over datasets that include ANLI (e.g., MNLI $+$ ANLI) is noteworthy, as ANLI's adversarial creation pipeline posed a much greater challenge for human workers, and used more existing resources to train model adversaries. 

Quite surprisingly, training on \wanli alone also outperforms combining \wanli with MultiNLI.
This reinforces that more data might not necessarily be better, especially when the data predominantly consists of easy-to-learn examples.


\begin{table}[t]
    \centering
    \resizebox{\columnwidth}{!}{%
    \begin{tabular}{lqswyz}
        \toprule
        &\multicolumn{5}{c}{\textbf{Test Set}}\\\cmidrule{2-6}
        & \multicolumn{1}{c}{\textbf{Diagnostics}} & \multicolumn{1}{c}{\textbf{HANS}*} &  \multicolumn{1}{c}{\textbf{ANLI}} & \multicolumn{1}{c}{\textbf{BIG-Bench}*} & \multicolumn{1}{c}{\textbf{\wanli}} \\\midrule
        ANLI & 65.67 & 80.58 & 55.21 & 77.10 & 63.85\\
        ANLI + \wanli & \textbf{72.82} & \textbf{88.58} & \textbf{56.59} & \textbf{84.89} & \textbf{75.84} \\
        \bottomrule
    \end{tabular}}
    \caption{
    Comparison of whether including \wanli in the training data of ANLI improves in-domain test performance, when finetuning RoBERTa-large.}
    \label{tab:results_id}
\end{table}

In addition to the OOD setting, we consider whether augmentation with \wanli can improve \textit{in-domain} test performance for another dataset (\autoref{tab:results_id}).
Indeed, augmenting ANLI's train set with \wanli improves test accuracy on ANLI by 1.4\%, while greatly aiding OOD test performance.
\section{Artifacts in \wanli}
\label{sec:analysis}

We next investigate whether \wanli contains similar artifacts to MultiNLI.\footnote{We note, however, that recent work has challenged whether artifacts based on partial input and lexical correlations in the dataset pose genuine robustness threats \cite{srikanth-rudinger-2022-partial, eisenstein-2022-informativeness}.} 
We find that while \wanli contains fewer previously known spurious correlations, it has a distinct set of lexical correlations that may reflect artifacts in GPT-3 output.

\subsection{Partial Input Models}
\label{subsec:partial-input}

Given that the task requires reasoning with both the premise and the hypothesis, a model that sees only one of the two inputs should have no information about the correct label.
We reproduce the methodology from \citet{gururangan-etal-2018-annotation} and train \texttt{fastText} classifiers to predict the label using partial input.
After first balancing \wanli, a model trained on just the hypotheses of \wanli achieves $41.6\%$ accuracy on the test set compared to $49.6\%$ for MultiNLI, when restricted to the same size. 
A premise-only model trained on \wanli achieves an accuracy of $42.9\%$.\footnote{Unlike \wanli, each MultiNLI premise is associated with hypotheses from all three labels; a premise-only baseline is thus guaranteed to have no information about the label.}

\subsection{Lexical Correlations}
\label{subsec:lexical-corr}

\begin{figure}
    \centering
    \includegraphics[width=\linewidth]{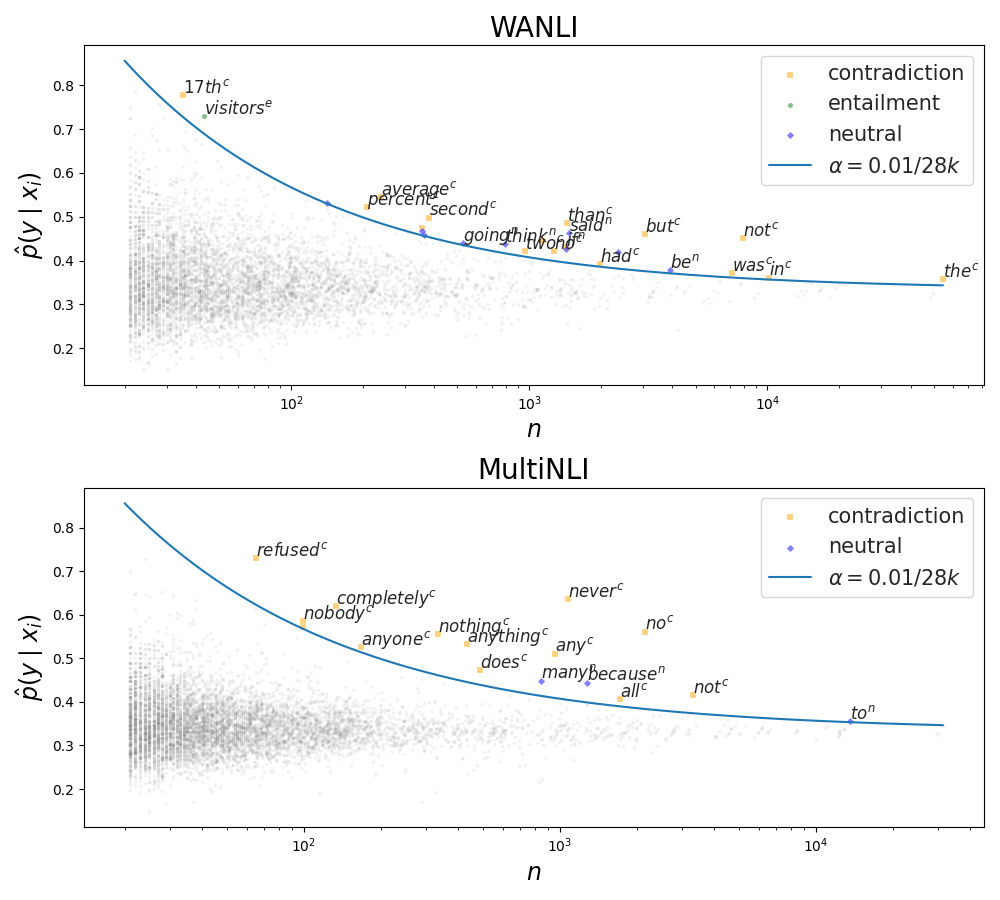}
    \caption{
    Competency problem-style statistical correlation plot between individual words and particular class labels, where the $y$-axis is the probability of label $y$ given the presence of the word $x_i$, and the $x$-axis is the number of times word $x_i$ appears in the data. 
    All points representing (word, label) pairs above the blue line have detectable correlations \cite{gardner-etal-2021-competency}.}
    \label{fig:competency_problems}
\end{figure}

\citet{gardner-etal-2021-competency} posit that all correlations between single words and output labels are spurious.
We plot the statistical correlation for every word and label in \autoref{fig:competency_problems}, after balancing \wanli and downsampling MultiNLI.
We observe that \wanli also contains words with detectable correlations, suggesting that GPT-3 may have some artifacts of its own due to the slightly different templates and different sets of in-context examples for each label.
Interestingly, the correlations tend to be a different set of words than for MultiNLI (other than ``not'' and ``no''), with less interpretable reasons for correlating with a certain label (e.g., ``second'', ``was'').

\subsection{Premise-Hypothesis Semantic Similarity}
\label{subsec:reasoning_diversity}

\begin{figure}[t]
    \centering
    \includegraphics[width=\linewidth]{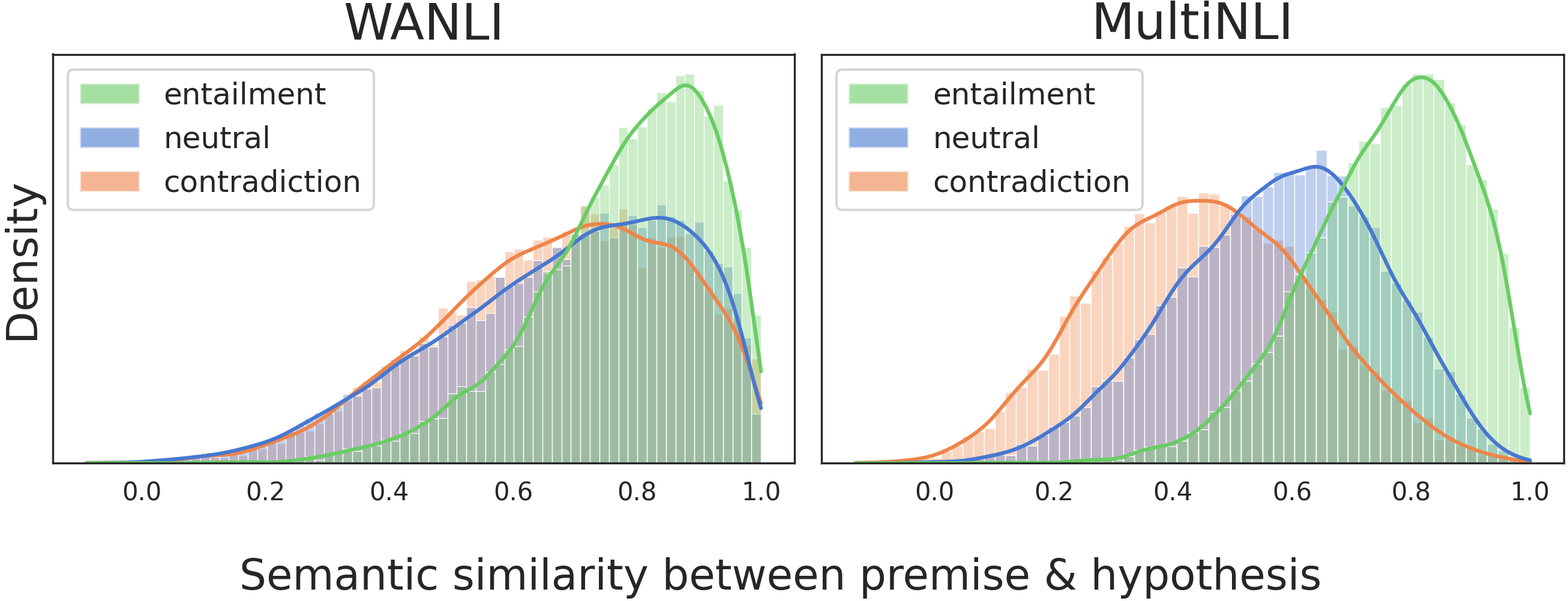}
    \caption{Semantic similarity between the premise and hypothesis, computed based on SBERT embeddings \cite{reimers-gurevych-2019-sentence}. 
    The distributions for each label class are much more well-separated in MultiNLI than in \wanli.}
    \label{fig:reasoning_diversity}
\end{figure}

We explore the semantic similarity between the premise and hypothesis within each label class using Sentence-BERT \cite{reimers-gurevych-2019-sentence}; these distributions are shown in \autoref{fig:reasoning_diversity}.
In both MultiNLI and \wanli, entailed hypotheses are naturally most semantically similar to the premise.
In MultiNLI, this is followed by neutral examples and then contradiction examples.
In contrast, in \wanli there is much greater overlap in the three distributions, and those for neutral and contradiction examples are nearly indistinguishable.
This suggests in \wanli, the semantic similarity between the premise and hypothesis provides less signal of the label.

\section{What does \wanli show about the human machine collaboration pipeline?}
\label{sec:discussion}
We discuss observations from collecting \wanli that may shed insight for future work in the direction of collaborative dataset creation.

\subsection{What kinds of revisions do annotators tend to make?}
\label{subsec:what-revisions?}

We find that revisions fall broadly into two categories: improving the fluency of the text, and improving the clarity of the relationship.
The majority of revisions change the length only slightly, with $74\%$ of both premise revisions and hypothesis revisions changing the word count between $-1$ and $+2$ words.
Fluency revisions often target well-documented issues with text generation, such as redundancy and self-contradiction.
Clarity revisions often resolve ambiguities in the example that make the entailment relationship difficult (or impossible) to determine, such as ambiguous coreference or temporal references.
We provide examples of revisions in Appendix \ref{subsec:revision_more}.

\begin{table*}[ht]
    \centering
    \resizebox{\textwidth}{!}{%
    \begin{tabular}{lll}
    \toprule
        \textbf{Example} & \textbf{Labels} & \textbf{Ambiguity} \\\midrule
        \makecell[l]{
        P: According to the most recent statistics, the rate of violent crime  \\
        in the United States has dropped by almost half since 1991. \\
        H: The rate of violent crime has not dropped by half since 1991.
        }
        & \makecell[l]{\colorbox{ent}{\textit{Entailment}}\\\colorbox{con}{\textit{Contradiction}}} & 
        \makecell[l]{Does ``almost half'' mean ``not \\
        half'' or ``basically half''?} \\\midrule
        \makecell[l]{
        P: As a result of the disaster, the city was rebuilt and it is now one\\
        of the most beautiful cities in the world.\\
        H: A disaster made the city better.}
        & \makecell[l]{\colorbox{ent}{\textit{Entailment}}\\\colorbox{neu}{\textit{Neutral}}}
        &\makecell[l]{Do indirect consequences\\ count?}\\\midrule
        \makecell[l]{
        P: It is a shame that the world has to suffer the pain of such \\unnecessary war.\\
        H: The world does not have to suffer such pain.
        }&\makecell[l]{\colorbox{ent}{\textit{Entailment}}\\\colorbox{con}{\textit{Contradiction}}}&
        \makecell[l]{
        Is the scope of ``has to''\\ in the hypothesis given \\ the war or not?}\\\midrule
        \makecell[l]{
        P: The original draft of the treaty included a clause that would have\\ 
        prohibited all weapons of mass destruction.\\
        H: The clause was removed in the final version of the treaty.}
        & \makecell[l]{\colorbox{ent}{\textit{Entailment}}\\\colorbox{neu}{\textit{Neutral}}} & 
        \makecell[l]{Does the premise imply that the \\
        clause is no longer in the treaty?}\\\midrule
        \makecell[l]{
        P: If you can't handle the heat, get out of the kitchen.\\
        H: If you can't handle the pressure, get out of the situation.}
        & \makecell[l]{\colorbox{ent}{\textit{Entailment}}\\\colorbox{neu}{\textit{Neutral}}} &
        \makecell[l]{Is the premise to be interpreted\\
        literally or figuratively?}\\\midrule
        \makecell[l]{
        P: In a world of increasing uncertainty, the only certainty is that \\
        nothing is certain.\\
        H: There is no certainty in the world.
        }&\makecell[l]{\colorbox{ent}{\textit{Entailment}}\\\colorbox{con}{\textit{Contradiction}}}&
        \makecell[l]{
        Self-contradictory but\\coherent premise}\\
    \bottomrule
    \end{tabular}}
    \caption{Examples where two annotators assigned different labels. We find that many examples represent genuinely ambiguous cases rather than careless mislabels, echoing previous findings \cite{pavlick-kwiatkowski-2019-inherent}.}
    \label{tab:disagreements}
\end{table*}
\subsection{What kinds of examples do annotators disagree on?}
\label{subsec:disagreement}

We find that examples on which annotators disagree provide an extremely interesting test bed for how ambiguities surface in classification tasks.
Upon inspecting the examples (some are shown in \autoref{tab:disagreements}), we observe that they represent genuinely ambiguous cases rather than careless mislabels, echoing previous findings \cite{pavlick-kwiatkowski-2019-inherent}. See further discussion in Appendix \ref{subsec:disagreement_more}.

\subsection{How reliably does GPT-3 reproduce the in-context pattern?}
\label{subsec:in-context}

One characteristic of \wanli is its imbalanced label distribution: even though the set of seed examples for generation was constructed to be balanced, after undergoing human labeling, only 15\% of examples are given the contradiction label.
We observe that contradiction patterns in in-context examples are generally much more challenging for GPT-3 to copy, likely because it was trained on (mostly) coherent sequences of sentences.
More broadly, we find that more abstract reasoning patterns are harder for GPT-3 to mimic than patterns that involve simpler transformations.

Nonetheless, even when GPT-3 does not successfully copy the examples, the diverse set of in-context examples leads to a variety of creative output that may be challenging for human crowdworkers to achieve.
\section{Related Work}
\label{sec:related}
\paragraph{Crowdsourcing} The scalability and flexibility of crowdsourcing has enabled the creation of foundational NLP benchmarks across a wide range of subproblems, and made it the dominant paradigm for data collection \cite[i.a.]{mihaylov-etal-2018-suit, rajpurkar-etal-2016-squad, huang-etal-2019-cosmos, talmor-etal-2019-commonsenseqa}. Nonetheless, a growing body of research shows that resulting datasets may not isolate the key linguistic phenomena \cite{jia-liang-2017-adversarial,chen-etal-2016-thorough,sugawara-etal-2019-assessing}.

For crowdsourcing NLI datasets, where the annotator is given a premise and asked to write a hypothesis of each label \cite{bowman-etal-2015-large, williams-etal-2018-broad}, the presence of annotation artifacts is especially well-studied \cite{gururangan-etal-2018-annotation,mccoy-etal-2019-right,glockner-etal-2018-breaking}.
Recent work attempted to remedy this through different data collection protocols but found negative results \cite{vania-etal-2020-asking,bowman-etal-2020-new}, showing this is a hard problem requiring greater innovation.

\paragraph{Adversarial data collection} In this paradigm, annotators are asked to produce examples on which current systems fail \cite[i.a.]{kiela-etal-2021-dynabench, talmor-etal-2021-commonsenseqa, zellers-etal-2019-hellaswag}. 
Beyond increasing annotator effort \cite{bartolo-etal-2020-beat}, adversarial methods have been challenged for not leading to better generalization on non-adversarial test sets \cite{kaushik-etal-2021-efficacy} and decreasing data diversity \cite{bowman-dahl-2021-will}.
Moreover, the resulting data has been shown to depend strongly on the adversaries, inhibiting a fair evaluation \cite{phang-etal-2021-adversarially}.
Finally, these approaches may produce examples beyond the scope of the task. 
For example, in Adversarial NLI \cite{nie-etal-2020-adversarial}, an estimated 58\% of examples required ``reasoning from outside knowledge or additional facts,'' which is arguably separate from the underlying problem of understanding semantic entailments.
We argue that we can better leverage the strengths of machines and humans by having them collaborate rather than act as adversaries.

\paragraph{Dataset generation}
Another recent approach leverages language models toward fully automatic dataset creation \cite[i.a.]{schick-schutze-2021-generating, wu-etal-2022-generating, west-etal-2021-symbolic, bartolo-etal-2021-improving}.
Removing human input may fundamentally limit the complexity of examples to phenomena already accessible by the model, when our goal is precisely to teach models more diverse phenomena.
The most similarly-motivated work to ours, \citet{lee-etal-2021-neural}, trains a data generator on ``data-rich slices'' of an existing dataset, and applies it to under-represented slices.
However, they use labels or metadata to represent slices, leaving automatic methods of identifying slices to future work.

\paragraph{Human-machine collaboration}
In terms of human-machine collaboration, \citet{tekiroglu-etal-2020-generating} and \citet{yuan-etal-2021-synthbio} employ a language model to generate counter-narratives to hate speech and biographies, respectively, which are validated and revised by humans. 
This was for a generative task, and we complement their findings by showing that human-machine collaboration can also be useful for generating labeled datasets for robust classification models.
Contemporary work \cite{bartolo-etal-2021-models} finetunes a generative annotation assistant to produce question-answer pairs that humans can revise for extractive QA. 

\section{Conclusion}
\label{sec:conclusion}

At the heart of dataset creation is distilling human linguistic competence into data that models can learn from.
The traditional crowdsourcing paradigm takes the view that the best approach for this is to solicit people to write free-form examples expressing their capabilities.
In this work, we present a worker-and-AI collaborative approach and apply it to create \wanli, whose empirical utility suggests that a better way of eliciting human intelligence at scale is to ask workers to \textit{revise} and \textit{evaluate} content.
To this end, we hope to encourage more work in developing generative algorithms to aid the dataset creation process, and therefore re-imagining the role of human annotation.


\section*{Acknowledgments}

We thank members of UW NLP, AI2, and Mila NLP for valuable feedback and discussion, and especially Jena Hwang for help in designing the AMT template, Julian Michael for countless discussions of NLI examples, and Alexander Fang for feedback during writing. We thank OpenAI for offering access to the GPT-3 API and the anonymous reviewers for valuable feedback.

This work was funded in part by the DARPA MCS program through NIWC Pacific (N66001-19-2-4031). The first author is supported by the National Science Foundation Graduate Research Fellowship Program.

\section{Ethics Statement}
\label{sec:ethics}
We acknowledge that text generated from large pretrained language models is susceptible to perpetuating social harms and containing toxic language \cite{sheng-etal-2019-woman, gehman-etal-2020-realtoxicityprompts}.
To partially remedy this, we ask annotators to discard any examples that may be perceived as offensive.
Nonetheless, it is possible that harmful examples (especially if they contain subtle biases) may have been missed by annotators and included in the final dataset.
Specifically due to the above harms, we additionally caution readers and practitioners against \textit{fully automating} any data creation pipeline.

In addition, we are cognizant of the asymmetrical relationship between requesters and workers in crowdsourcing. 
We took great care to pay fair wages, and were responsive to feedback and questions throughout the data collection process (see \autoref{sec:human_annotation} for details).
The only personal information we collect is the worker IDs from Amazon Mechanical Turk, which we will not release.
The annotation effort received an IRB exemption.
\section{Limitations}
\label{sec:limitations}

In this paper, we apply our collaborative dataset creation pipeline to a single language and task, English natural language inference, and leave application of the pipeline more broadly to future work. 

It is possible (if not likely) that datasets partially authored by language models will have artifacts of their own, especially those reflecting social biases that may not be captured by our accuracy-based evaluation setup. 
For investigation of a specific generation artifact observed by \citet{yuan-etal-2021-synthbio} in their own collaborative dataset, namely the over-representation of Western entities, please see Appendix \ref{subsec:western_entities}.

We are not able to perform ablations on different parts of the pipeline to understand the effectiveness of each component, e.g., by comparing different means of collecting exemplar groups or different templates for prompting GPT-3.
Unfortunately, such variations would be prohibitively expensive as they each require collecting a dataset of sufficient scale (along with the necessary human annotation). 

Finally, although we uncover examples where annotators disagree for valid reasons (see \autoref{tab:disagreements}), we only use one label per example for training and evaluation. 
This is because to show the effectiveness of \wanli, we need to compare \wanli to existing (singly-labeled) training datasets via performance on established (singly-labeled) benchmarks. 
We encourage future work to understand the limitations of forcing inherently ambiguous instances into the $n$-way classification scheme, or otherwise discarding these potentially valuable examples of linguistic reasoning as noise.

\bibliography{anthology,custom}
\bibliographystyle{acl_natbib}
\appendix

\section{Estimated Max Variability}\label{sec:max_variability}
In order to test the correlation between variability and estimated max variability on a dataset $\mathcal D$, we would have to repeatedly hold out a single example $x$, train a model on $\mathcal D\setminus\{x\}$, and evaluate how well the estimated max variability from the model trained on $\mathcal D\setminus\{x\}$ correlates with the true variability from the model trained on $\mathcal D$, which saw $x$ during training.

\begin{figure}[ht]
    \centering
    \includegraphics[scale=0.5]{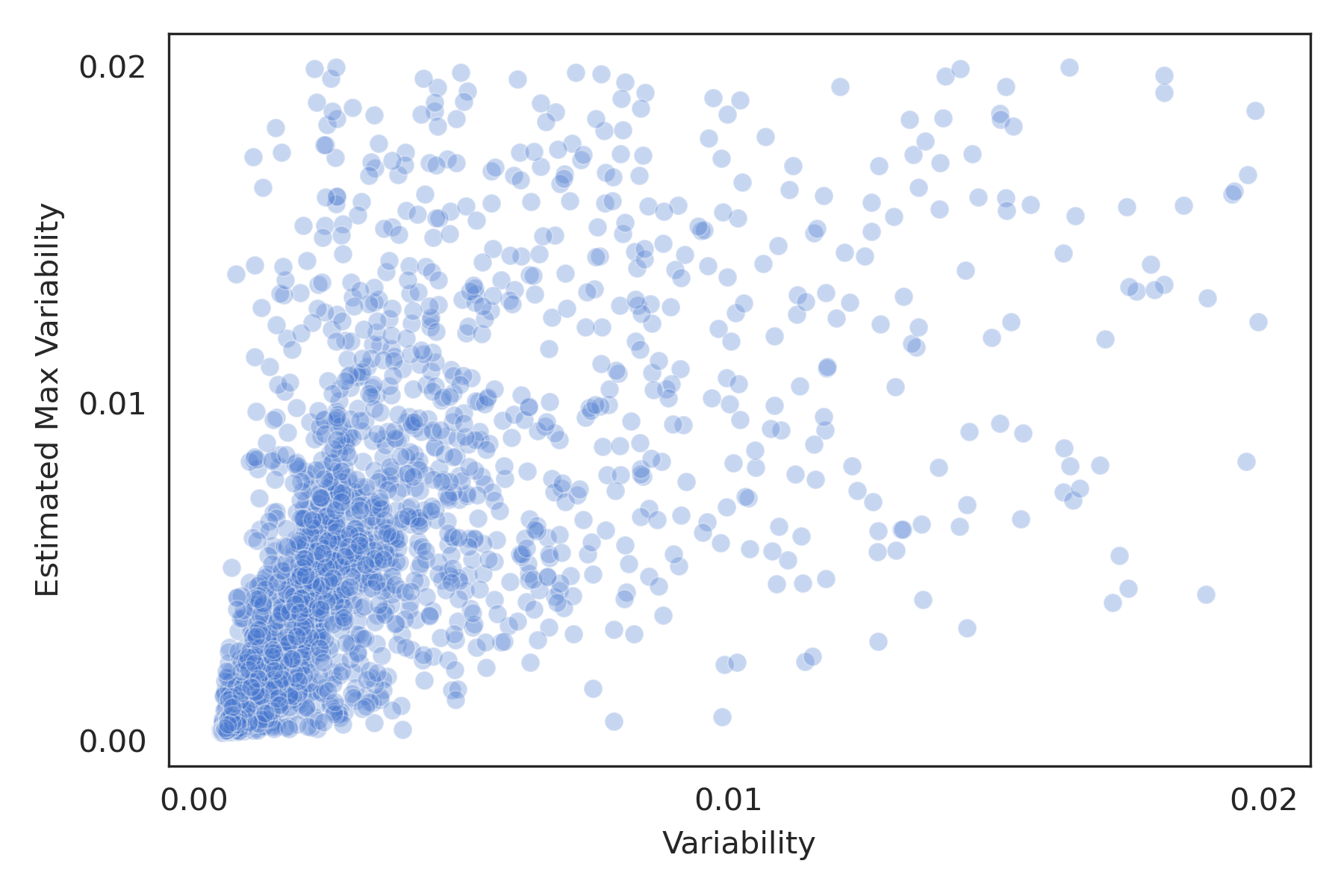}
    \caption{Correlation between variability of examples on a model that trains on the full MNLI dataset, and estimated max variability of the same examples when they are held out of the training set.}
    \label{fig:max_variability}
\end{figure}

Unfortunately, this would be a very expensive experiment. Instead, we split the MNLI train set into $99\%$ for training and $1\%$ (3928 examples) for evaluation. For each of the held-out examples, we calculate the variability under $\mathcal M\textsubscript{MNLI}$ and estimated max variability under $\mathcal M\textsubscript{MNLI 99\%}$. The correlation is shown in \autoref{fig:max_variability}, and has a Pearson's correlation coefficient of 0.527 with a $p$-value of $7 \times 10^{-281}$.

\section{Modeling Details}\label{sec:modeling_details}
All model training is implemented with the HuggingFace \cite{wolf-etal-2020-transformers} library and uses the original hyperparameters from the RoBERTa paper for finetuning on GLUE \cite{liu-etal-2019-roberta}. 
We train the model for five epochs and evaluate the final model.
We choose not to use an early stopping scheme in order to isolate the training data as the object of study and control for training length as a confounding factor. This is important since \citet{tu-etal-2020-empirical} showed that counter-examples can be learned better with longer training.

All training was performed on a single Nvidia Quadro RTX 6000 GPU. The duration of training varied depending on the size of the training data, from 3 hours for \wanli to 14 hours for MultiNLI $+$ \wanli.

\begin{table}[ht]
    \centering
    \begin{tabular}{cc}
    \toprule
        \textbf{Hyperparameter} & \textbf{Assignment} \\\midrule
        Model & RoBERTa-large\\
        Number of parameters & 345M \\
        Number of epochs & 5 \\
        Learning rate & 10$^{-5}$ \\
        Batch size & 32 \\
        Weight decay & 0.1 \\
        Learning rate decay & linear \\
        Warmup ratio & 0.06 \\
    \bottomrule
    \end{tabular}
    \caption{Training hyperparameters for RoBERTa-large.}
    \label{tab:training_hyperparameters}
\end{table}
\section{\wanli Details and Discussion}\label{sec:dataset_details}

\subsection{Example GPT-3 Context}\label{subsec:context_examples}
We include some examples of full GPT-3 contexts in \autoref{tab:row1}, \ref{tab:row2}, \ref{tab:row3}, \ref{tab:row7}.

\subsection{GPT-3 Generation Hyperparameters}\label{subsec:generation_hyperparams}

We queried the GPT-3 Curie model available through the OpenAI API\footnote{\url{https://openai.com/api}} on the dates November 3 to November 5, 2021.
In total, the generation cost \$677.89.
Hyperparameters for generation\footnote{described at \url{https://beta.openai.com/docs/api-reference/completions/create}} are shown in \autoref{tab:generation_hyperparameters}. 

\begin{table}[ht]
    \centering
    \begin{tabular}{cc}
    \toprule
        \textbf{Hyperparameter} & \textbf{Assignment} \\\midrule
        Top $p$ & 0.5 \\
        Temperature & 1 \\
        Max tokens & 120 \\
        Stop string & \texttt{\textbackslash n\textbackslash n}\\
        Presence penalty & 0.0 \\
        Frequency penalty & 0.0 \\
    \bottomrule
    \end{tabular}
    \caption{Hyperparameters for generation from GPT-3.}
    \label{tab:generation_hyperparameters}
\end{table}

\subsection{Dataset sizes at each stage}\label{subsec:dataset_sizes}
In Stage 1, we collect the top 25\% most ambiguous examples from each label class in MultiNLI as our set of seed examples. This leads to 98,176 seed examples, where each seed example corresponds to a unique context for GPT-3.
We generate $n=5$ examples per seed example, and skip examples that are not properly formatted with a distinct premise and hypothesis following the context template (\autoref{fig:template}).
At the end of Stage 2, the size of $\mathcal D\textsubscript{gen}$ is 372,404.
After applying the filtering heuristics described in \S\ref{subsec:filtering} on $\mathcal D\textsubscript{gen}$, the remaining dataset size is 287,241.
Of the examples discarded, 79,278 generated examples had identical premise and hypothesis (sans punctuation and casing), and 4,732 examples had copied an in-context example.
Next, we keep the half with the highest estimated max variability by sourcing an equal number of examples from each (intended) label class for a balanced dataset, resulting in $\mathcal D\textsubscript{filtered}$ with size 143,619.
However, we do not actually recruit human review on all of $\mathcal D\textsubscript{filtered}$, and instead annotate a total of 118,724 examples.
Since some of these examples are discarded, the final \wanli dataset contains 107,885 examples.
These correspond to 57,825 seed examples from MultiNLI.

\subsection{Investigation of Western entities in \wanli versus MNLI}\label{subsec:western_entities}
While we investigated known artifacts of crowdsourced datasets in \S\ref{sec:analysis}, generated datasets may have distinct kinds of artifacts.
Indeed, recent related work qualitatively observed an over-representation of Western entities in generated biographies \cite{yuan-etal-2021-synthbio}.
To investigate whether this is also characteristic of \wanli, we use \texttt{flair} \cite{akbik-etal-2019-flair} to perform named entity recognition on MultiNLI and \wanli.
Due to the challenges and ethical risks of automatically determining the origin of names and organizations, we focus on the diversity of locations mentioned.
We use \texttt{geopy}\footnote{\url{https://geopy.readthedocs.io}} to map all locations (e.g., cities, provinces, landmarks, as well as countries) to a country.

We find that 79\% of location mentions in \wanli are in Europe or North America, compared to 71\% in MultiNLI.
In particular, the United States is massively over-represented, accounting for 46\% of mentions in \wanli and 26\% in MultiNLI.
However, both datasets feature a diversity of location names: \wanli mentions locations in 210 countries across 22K location entities, and MultiNLI mentions locations in 227 countries across 163K location entities.
We conclude that over-representation of Western entities is indeed a concern for generated datasets, and encourage future work to consider this.
\section{Human Review}
\label{sec:human_annotation}
\begin{figure*}
    \centering
    \includegraphics[width=0.9\textwidth]{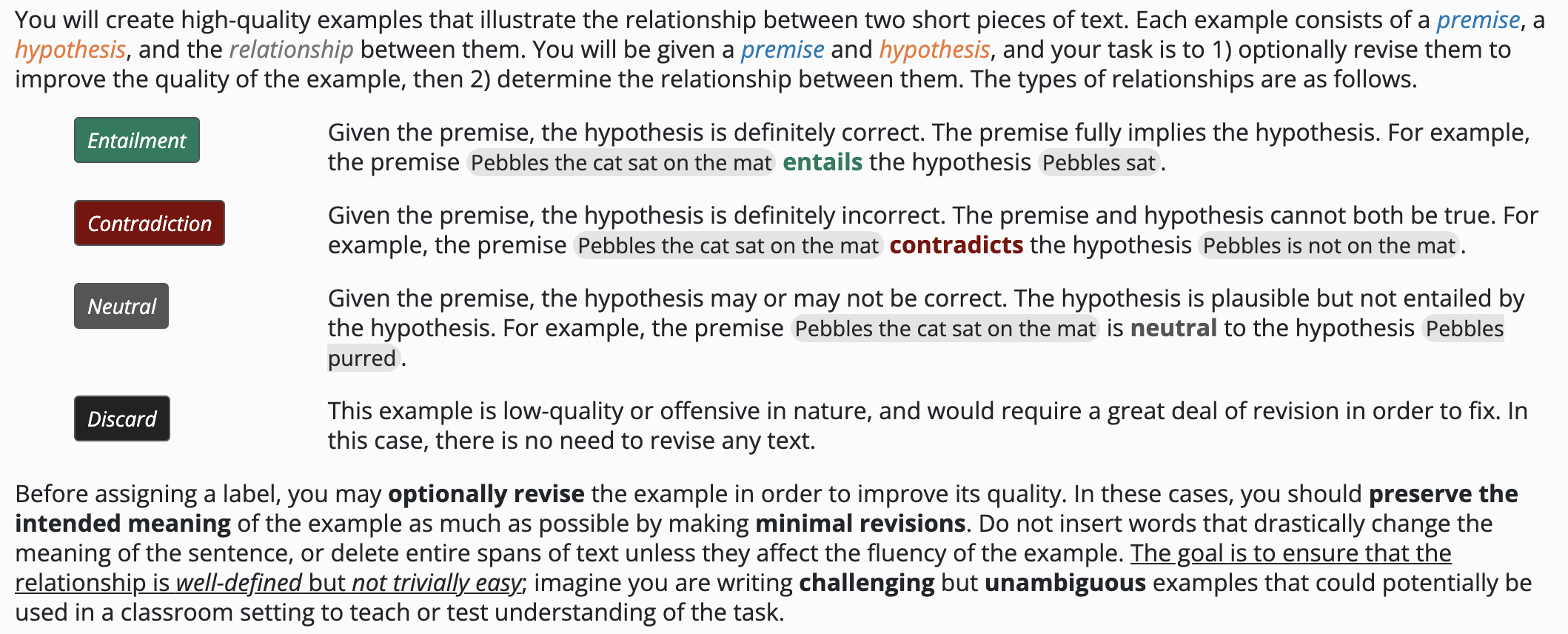}
    \caption{Instructions provided to crowdworkers on Amazon Mechanical Turk.}
    \label{fig:amt_instruction}
\end{figure*}

\begin{figure*}
    \centering
    \includegraphics[width=\textwidth]{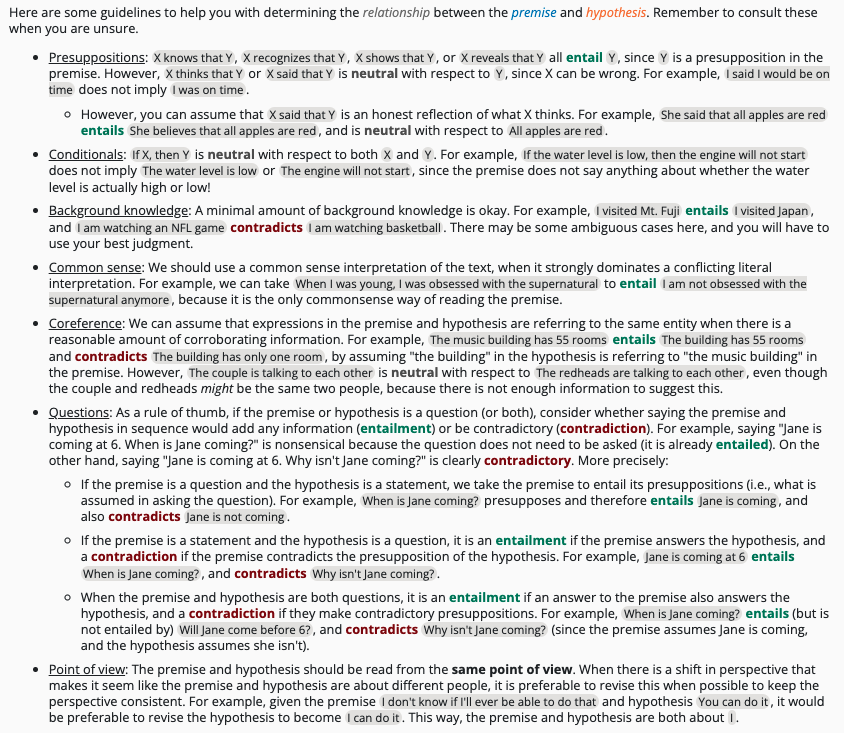}
    \caption{Guidelines provided to crowdworkers in the human review stage.}
    \label{fig:amt_guidelines}
\end{figure*}

\begin{figure*}
    \centering
    \includegraphics[width=0.7\textwidth]{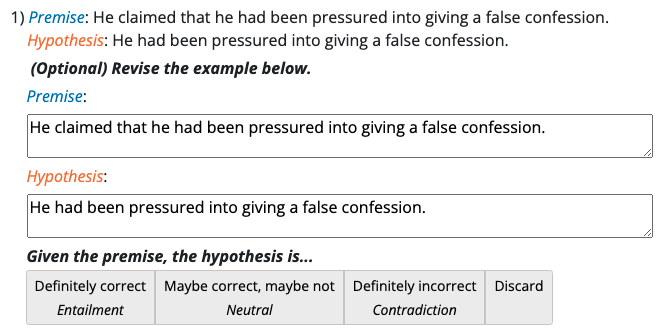}
    \caption{The interface on Amazon Mechanical Turk used for collecting human annotations. Annotators are given free text boxes that are pre-populated with the original premise and hypothesis, to ease the work of revision. Then, they either select an entailment class or discard the example.}
    \label{fig:amt_interface}
\end{figure*}

Screenshots of the instructions, guidelines, and annotation interface are shown in Tables \ref{fig:amt_instruction}, \ref{fig:amt_guidelines}, and \ref{fig:amt_interface}. 
The guidelines take inspiration from the design of the NLI Diagnostics dataset \cite{wang-etal-2018-glue}.
To collect a pool of qualified workers, we designed a qualification task with examples testing each of these categories.
NLI is a challenging task, and many generated examples are especially challenging by design.
Therefore, instructing annotators in how to think about the task and resolve common issues is key to collecting high-quality, label-consistent data.

\subsection{The Annotators}\label{subsec:annotators}
Annotators were required to have a HIT approval rate of 98\%, a total of 10,000 approved HITs, and be located in the United States.

300 Turkers took our qualification test, of which 69 passed.
Turkers who were later found to produce extremely careless annotations were removed from the qualification list (and oftentimes, their annotations were discarded, though they were still paid for their work).
The number of workers who contributed to the final dataset is 62.

Throughout the data collection process, the authors would review annotations and write individualized emails to Turkers with feedback, as well as group emails to clarify common challenging cases of NLI (such as examples involving questions).
This follows the recommended crowdsourcing protocol from \citet{nangia-etal-2021-ingredients}.

\subsection{Compensation}
In designing the task, we aimed for a pay rate of at least \$15 per hour.
Workers were paid \$0.12 for each example that they annotate.
At the end of data collection, we aggregate the earning and time spent from each crowdworker, and find that the median hourly rate was \$22.72, with 85\% of workers being paid over the \$15/hour target.

\subsection{Revision Analysis}\label{subsec:revision_more}

We provide examples of revisions in \autoref{tab:revisions}. 
We find that revisions are generally targeted yet effective. 
The majority of revisions change the length only slightly, with $74\%$ of both premise revisions and hypothesis revisions changing the word count between $-1$ and $+2$ words. 
A very large proportion, 11.6\% of premise revisions and 20.6\% of hypothesis revisions, changed the set of pronouns present in the text, often to clarify coreference.

We instructed annotators to revise examples only when it would make the example more ``interesting'' in some sense, or more clear without removing what's interesting.
Nonetheless, we still observed a large number of revisions that greatly simplified the example, oftentimes re-introducing the same artifacts that have been documented in prior work.
Therefore, we ultimately chose to include revisions only when both annotators revised the example, indicating that the revision was necessary to improve the quality of the example.

\subsection{Disagreement Analysis}\label{subsec:disagreement_more}
In order to investigate the utility of collecting a third annotation, we randomly sampled 80 examples where the two annotators disagreed on the label (and neither revised nor discarded), and two of the authors separately annotated each one.
Shockingly, the two authors agreed on the label only 49\% of the time.
Furthermore, in 12\% of cases, all three labels were present among the four annotations.
This suggests that disagreement is often due to true ambiguity rather than careless mislabeling, and a third annotation would be unlikely to have high payoff in terms of ``correcting'' the label.
As a result, we choose not to collect a third annotation in this work.
Instead, we believe that the doubly-annotated examples in \wanli have flagged many interesting cases of ambiguity in NLI, and we encourage future work to design richer annotation frameworks to uncover the source(s) of ambiguity.

We choose to keep examples with disagreement in the \wanli dataset because we believe that finetuning with one of multiple reasonable labels still provides valuable training signal.

\begin{table}[ht]
    \centering
    \resizebox{\columnwidth}{!}{%
    \begin{tabular}{clcc}
        \toprule
        &&\multicolumn{2}{c}{\textbf{MNLI Dev.~Set}}\\\cmidrule{3-4}
        && Matched & Mismatched \\\midrule
        \multirow{4}{*}{\rotatebox[origin=c]{90}{\textbf{Train Set}}} & MNLI & 90.30 & 90.10 \\
        & MNLI $\diamond$ \wanli & 89.63 & 88.95 \\
        & MNLI + \wanli & 89.90 & 89.32 \\
        & \wanli & 80.17 & 80.46 \\
        \bottomrule
    \end{tabular}}
    \caption{Results on MultiNLI's development set.}
    \label{tab:mnli_results}
\end{table}

\begin{table*}[ht]
    \centering
    \resizebox{\textwidth}{!}{%
    \begin{tabular}{lll}
    \toprule
        \textbf{Example} & \textbf{Label} & \textbf{Purpose of Revision} \\\midrule
        \makecell[l]{
        P: \colorbox{insert}{\textcolor{insertt}{The power plant}} \colorbox{delete}{\textcolor{deletee}{\sout{It}}} is the only source of continuous electric power for\\
        the city.\\
        H: The power plant is very important for the city.
        }&\colorbox{ent}{\textit{Entailment}}&\makecell[l]{Coreference\\resolution}\\\midrule
        \makecell[l]{
        P: It was a well-known fact that \colorbox{delete}{\textcolor{deletee}{\sout{it was a well-known fact that}}} the solution \\
        was well-known.\\
        H: The solution was well-known.
        }& \colorbox{ent}{\textit{Entailment}} & Redundancy\\\midrule
        \makecell[l]{
        P: This will be the first time the king has met the queen in person.\\
        H: The king has met the queen \colorbox{insert}{\textcolor{insertt}{in person}} before.
        }&\colorbox{con}{\textit{Contradiction}}&Clarity\\\midrule
        \makecell[l]{
        P: She walked with a light step, as if she were floating on air.\\
        H: She was floating on air\colorbox{delete}{\textcolor{deletee}{\sout{, as if she were walking on air}}}.}
        & \colorbox{con}{\textit{Contradiction}} & Coherence\\\midrule
        \makecell[l]{
        P: There is a slight possibility that, if the same temperature data are used, \\
        the temperature of the Earth's surface in 1998 will be lower than the \\
        temperature of the Earth's surface \colorbox{delete}{\textcolor{deletee}{\sout{in 1998}}} \colorbox{insert}{\textcolor{insertt}{now}}.\\
        H: The Earth's surface in 1998 was lower than the Earth's surface \colorbox{delete}{\textcolor{deletee}{\sout{in 1998}}}\\ \colorbox{insert}{\textcolor{insertt}{now}}.
        }& \colorbox{neu}{\textit{Neutral}} & Self-contradiction\\\midrule
        \makecell[l]{
        P: She had to go to the library to find out what the name of the street was.\\
        H: She \colorbox{insert}{\textcolor{insertt}{already}} knew the name of the street.}
        &\colorbox{con}{\textit{Contradiction}}
        &\makecell[l]{
        Ambiguous temporal \\ reference}\\\midrule
        \makecell[l]{
        P: A number of theories have been proposed to explain the decline of \\
        violence in modern society.\\
        H: Violence \colorbox{delete}{\textcolor{deletee}{\sout{will decline}}}\colorbox{insert}{\textcolor{insertt}{has declined}} in modern society.
        }&\colorbox{ent}{\textit{Entailment}}&Consistent tense\\
    \bottomrule
    \end{tabular}}
    \caption{Some examples of revisions that were done by annotators on examples generated by GPT-3.}
    \label{tab:revisions}
\end{table*}
\section{Additional Experiments}
\label{sec:additional_experiments}

\subsection{Additional baselines}\label{subsec:additional_baselines}
We additionally perform comparisons with several subsets of MultiNLI which are the same size as \wanli: MultiNLI filtered with the AFLite algorithm (MultiNLI with AFLite; \citealp{lebras-etal-2020-adversarial}), the most ambiguous examples of MultiNLI (MultiNLI ambiguous; \citealp{swayamdipta-etal-2020-dataset}), and a random subset of MultiNLI (MultiNLI downsampled).
Results in \autoref{tab:additional_results} show that a \wanli-trained model outperforms these baselines on every test set.

\subsection{Evaluation on MultiNLI}\label{subsec:mnli_results}
We report the results on MultiNLI's development set in \autoref{tab:mnli_results}. We find that mixing \wanli into the MultiNLI training data (either through swapping or augmentation) maintains in-domain accuracy within $\sim$1\%. Training on \wanli alone drops performance on MultiNLI's development set by $\sim$10\%; however, the higher performance on other out-of-domain test sets suggests that evaluation through MultiNLI may not be a definitive signal of model ability.

\subsection{Finetuning T5}\label{subsec:finetuning_t5}
We demonstrate that the robustness improvements from training on \wanli generalizes to another model architecture, T5-base \cite{raffel-etal-2020-exploring}, which was never used in the data curation pipeline.
Shown in \autoref{tab:t5_results}, training T5-base on \wanli also outperforms training on MultiNLI on every test set, including by 4\% of NLI Diagnostics, 10\% on HANS, and 8\% on Adversarial NLI (similar margins compared to finetuning RoBERTa-large).
\section{Data Map of \wanli}
\begin{figure*}
    \centering
    \begin{subfigure}{.5\textwidth}
      \centering
      \includegraphics[width=\linewidth]{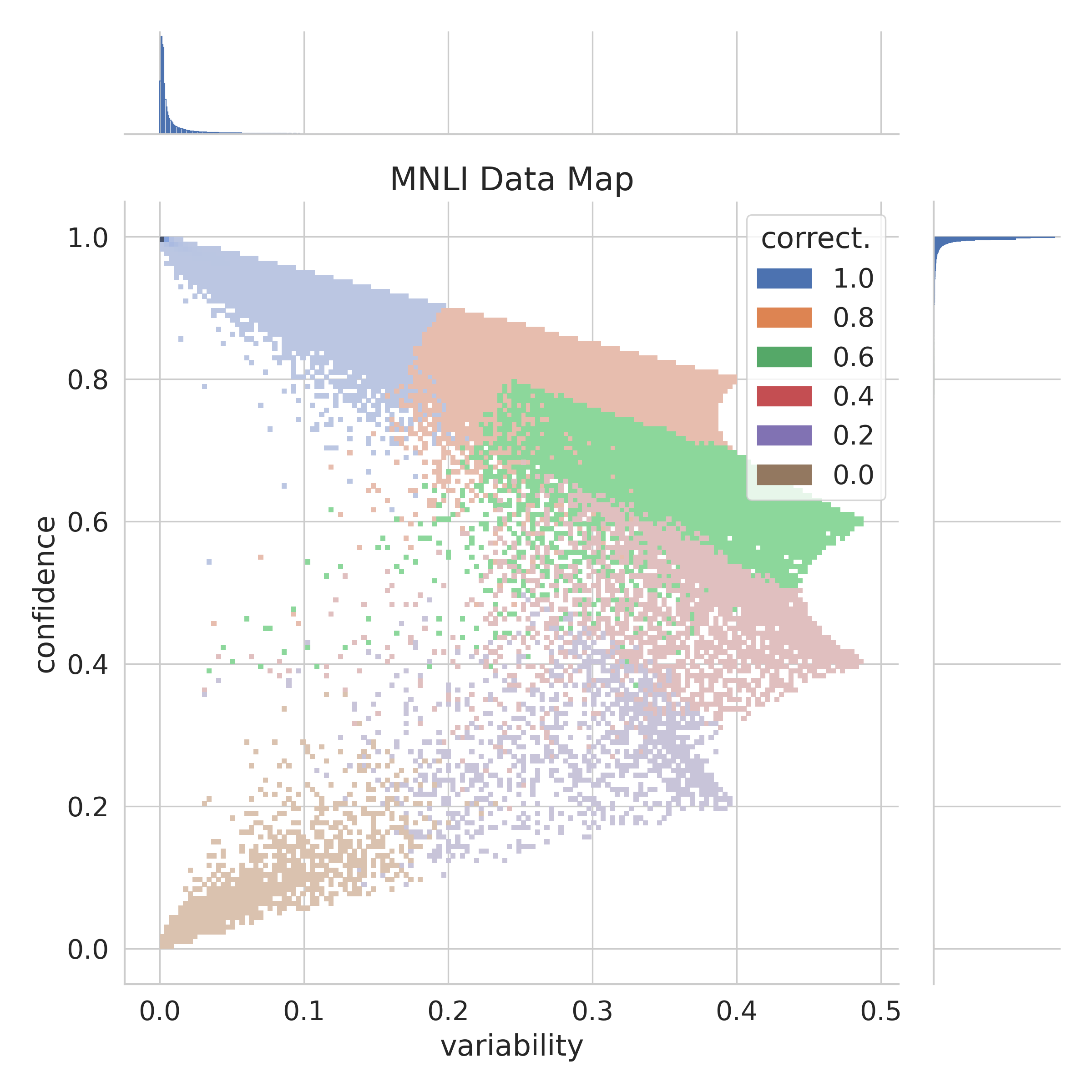}
    \end{subfigure}%
    \begin{subfigure}{.5\textwidth}
      \centering
      \includegraphics[width=\linewidth]{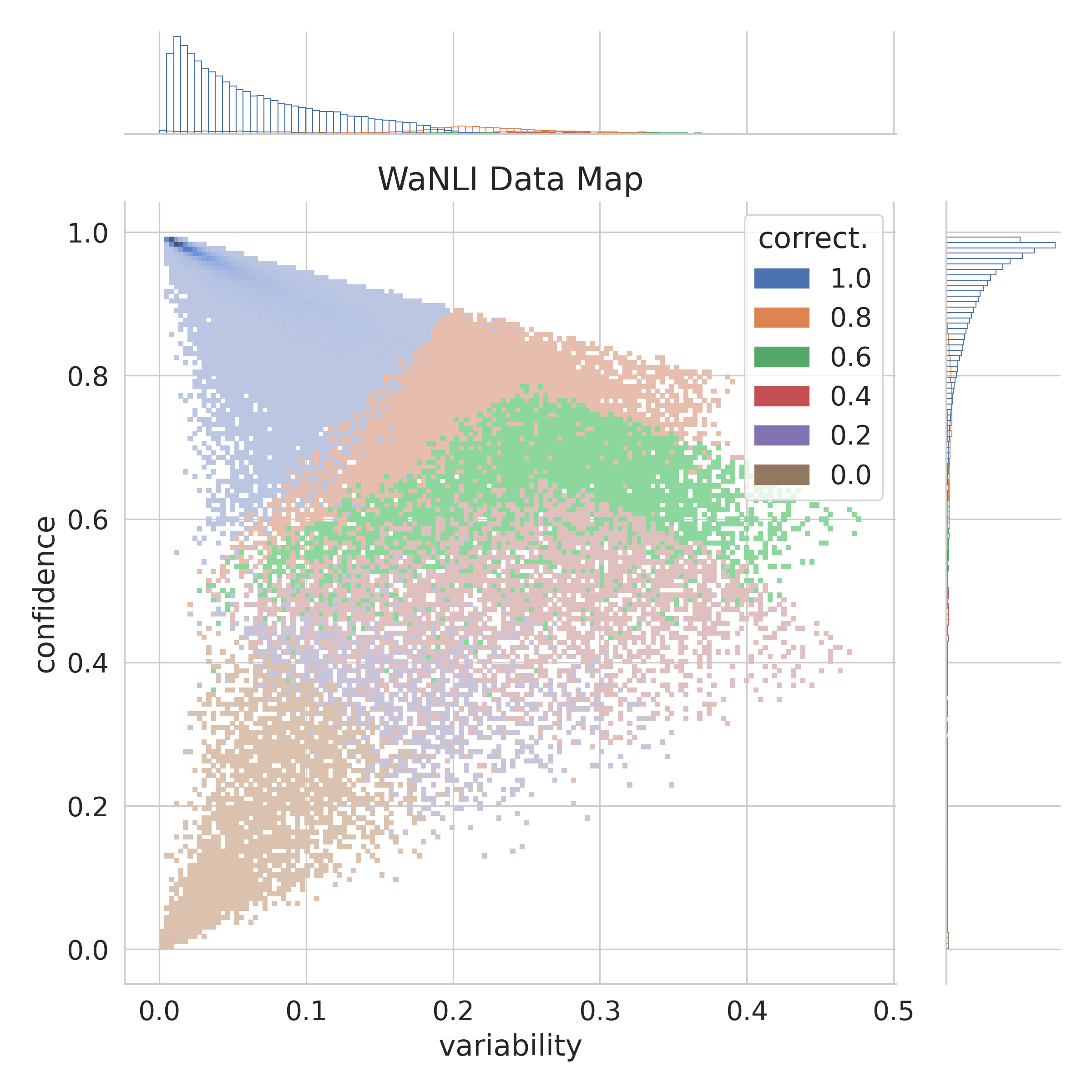}
    \end{subfigure}
    \caption{\textbf{Left}: Data map for MultiNLI train set, based on a RoBERTa-large classifier trained on MultiNLI. \textbf{Right}: Data map for \wanli train set, based on a RoBERTa-large classifier trained on \wanli. A comparison of the distribution in variability (which determines example ambiguity) is remarkable -- we see that MultiNLI is overwhelmingly dominated by easy-to-learn examples with variability close to 0. In contrast, the distribution in variability is much more spread out in \wanli, suggesting that the dataset contains more valuable examples overall.}
    \label{fig:data_maps}
\end{figure*}

In \autoref{fig:data_maps}, we show a data map of MultiNLI relative to RoBERTa-large trained on MNLI, and of \wanli relative to RoBERTa-large trained on \wanli. 

\begin{table*}
    \centering
    \resizebox{\textwidth}{!}{%
    \begin{tabular}{clcqstuvwxyz}
        \toprule
        &&&\multicolumn{9}{c}{\textbf{Test Set}}\\\cmidrule{4-12}
        &&&\multicolumn{1}{c}{\textbf{Diagnostics}} & \multicolumn{1}{c}{\textbf{HANS}*} & \multicolumn{1}{c}{\textbf{QNLI}*} & \multicolumn{1}{c}{\textbf{WNLI}*} & \multicolumn{1}{c}{\textbf{NQ-NLI}*} & \multicolumn{1}{c}{\textbf{ANLI}} & \multicolumn{1}{c}{\textbf{FEVER-NLI}} &  \multicolumn{1}{c}{\textbf{BIG-Bench}*} & \multicolumn{1}{c}{\textbf{\wanli}} \\
        &&\multicolumn{1}{r}{Data size}& 1104 & 30K & 5266 & 706 & 4855 & 3200 & 20K & 3324 & 5000 \\\midrule
        \multirow{6}{*}{\rotatebox[origin=c]{90}{\textbf{Training Set}}} & MNLI & 393K & 68.47 & 78.08 & 52.69 & 56.09 & 62.34 & 32.37 & 68.29 & 64.68 & 64.62\\\cmidrule{2-12}
        & MNLI (AFLite) & 103K & 60.50 & 73.73 & 53.91 & 56.37 & 64.28 & 33.12 & 68.04 & 70.75 & 62.19 \\
        & MNLI (ambiguous) & 103K & 65.03 & 74.93 & 54.42 & 62.32 & 62.14 & 32.68 & 67.42 & 68.77 & 61.15 \\
        & MNLI (downsampled) & 103K & 64.67 & 71.15 & 59.15 & 52.97 & 62.14 & 28.99 & 69.08 & 56.76 & 62.84\\\cmidrule{2-12}
        & \textbf{\wanli} & 103K & \textbf{72.55} & \textbf{89.40} & \textbf{76.81} & \textbf{65.15} & \textbf{64.03} & \textbf{41.12} & \textbf{70.63} & \textbf{75.40} & \cellcolor[gray]{0.8}{75.49}\\
        \bottomrule
    \end{tabular}}
    \caption{Additional baselines that finetune RoBERTa-large on different subsets of MultiNLI, filtered via existing debiasing methods.}
    \label{tab:additional_results}
\end{table*}
\begin{table*}
    \centering
    \resizebox{\textwidth}{!}{%
    \begin{tabular}{clcqstuvwxyz}
        \toprule
        &&&\multicolumn{9}{c}{\textbf{Test Set}}\\\cmidrule{4-12}
        &&&\multicolumn{1}{c}{\textbf{Diagnostics}} & \multicolumn{1}{c}{\textbf{HANS}*} & \multicolumn{1}{c}{\textbf{QNLI}*} & \multicolumn{1}{c}{\textbf{WNLI}*} & \multicolumn{1}{c}{\textbf{NQ-NLI}*} & \multicolumn{1}{c}{\textbf{ANLI}} & \multicolumn{1}{c}{\textbf{FEVER-NLI}} &  \multicolumn{1}{c}{\textbf{BIG-Bench}*} & \multicolumn{1}{c}{\textbf{\wanli}} \\
        &&\multicolumn{1}{r}{Data size}& 1104 & 30K & 5266 & 706 & 4855 & 3200 & 20K & 3324 & 5000 \\\midrule
        \multirow{5}{*}{\rotatebox[origin=c]{90}{\textbf{Training Set}}} & MNLI & 393K & 60.87 & 76.40 & 65.49 & 50.56 & 61.33 & 30.56 & 66.94 & 58.87 & 61.72 \\\cmidrule{2-12}
        & MNLI + Tailor & 485K & 61.14 & 74.34 & 63.33 & 50.70 & 62.05 & 31.06 & 67.15 & 68.95 & 61.28 \\
        & MNLI + Z-Aug & 754K & 60.05 & 76.73 & 63.46 & 50.14 & 60.53 & 32.50 & 67.10 & 54.81 & 61.38 \\
        & MNLI $\diamond$ ANLI & 393K & 61.23 & 73.55 & 69.80 & 52.26 & 61.64 & \cellcolor[gray]{0.8}{49.91} & \cellcolor[gray]{0.8}{70.82} & 68.80 & 61.66 \\\cmidrule{2-12}
        & \textbf{\wanli} & 103K & \textbf{64.58} & \textbf{86.25} & \textbf{74.66} & \textbf{51.13} & \textbf{63.66} & \textbf{38.22} & \textbf{68.27} & \textbf{76.17} & \cellcolor[gray]{0.8}{72.56} \\
        \bottomrule
    \end{tabular}}
    \caption{Empirical comparison of different training datasets for T5-base. For brevity, we include MNLI, \wanli, and the strongest baselines from the results based on RoBERTa-large from \autoref{tab:results}.}
    \label{tab:t5_results}
\end{table*}

\begin{table*}[t]
    \centering
    \resizebox{\textwidth}{!}{%
    \begin{tabular}{l}
    \toprule
    \makecell[l]{
    Write a pair of sentences that have the same relationship as the previous examples. Examples:\\[4pt]
    1. In \textbf{\textit{six states}}, the federal investment represents almost the entire contribution for providing civil legal services\\ to low-income individuals.\\
    Implication: In \textbf{\textit{44 states}}, the federal investment does not represent the entire contribution for providing civil \\
    legal services for people of low income levels.\\[4pt]
    2. But if it's at all possible, plan your visit for the \textbf{\textit{spring, autumn, or even the winter}}, when the big sightseeing\\ 
    destinations are far less crowded.\\
    Implication: This destination is most crowded in the \textbf{\textit{summer}}. \\[4pt]
    3. \textbf{\textit{5 percent}} of the routes operating at a loss.\\
    Implication: \textbf{\textit{95 percent}} of routes are operating at either profit or break-even.\\[4pt]
    4. 30 About \textbf{\textit{10 percent}} of households did not\\
    Implication: Roughly \textit{\textbf{ninety percent}} of households did this thing.\\[4pt]
    5. \textbf{\textit{5 percent}} probability that each part will be defect free.\\
    Implication: Each part has a \textbf{\textit{95 percent}} chance of having a defect.\\[4pt]
    6.}\\
    \bottomrule
    \end{tabular}}
    \caption{Context corresponding to row 1 in \autoref{tab:examples}, which contains \colorbox{ent}{\textit{Entailment}} examples from MultiNLI found via nearest neighbors in \texttt{[CLS]} token embedding space.
    All examples require reasoning about set complements, including from the universe of 100 percent, the 50 U.S. states, as well as the four seasons.}
    \label{tab:row1}
\end{table*}
\begin{table*}[b]
    \centering
    \resizebox{\textwidth}{!}{%
    \begin{tabular}{l}
    \toprule
    \makecell[l]{
    Write a pair of sentences that have the same relationship as the previous examples. Examples:\\[4pt]
    1. Small holdings abound, and traditional houses sit low on the treeless hillsides.\\
    Possibility: The hills were \textbf{\textit{the only place}} suitable to build traditional houses.\\[4pt]
    2. The inner courtyard has a lovely green and blue mosaic of Neptune with his wife Amphitrite.\\
    Possibility: \textbf{\textit{The only colors}} used in the mosaic of Neptune and Amphitrite are green and blue. \\[4pt]
    3. Nathan Road, Central, and the hotel malls are places to look.\\
    Possibility: \textbf{\textit{The only places}} to look are Nathan Road, Central and hotel malls.\\[4pt]
    4. Make your way westward to the Pont Saint-Martin for a first view of the city's most enchanting quarter, \\
    the old tannery district known as Petite France.\\
    Possibility: \textbf{\textit{The only place}} to the west of Pont Saint-Martin is the old tannery district.\\[4pt]
    5. The artisans, tradespeople, and providers of entertainment (reputable and not so reputable) lived downtown \\
    on the reclaimed marshlands north and east, in the area still known as Shitamachi.\\
    Possibility: \textbf{\textit{The only place}} where artisans, tradespeople and entertainers could live was in the marshlands to\\
    the north and east.\\[4pt]
    6.}\\
    \bottomrule
    \end{tabular}}
    \caption{Context corresponding to row 2 in \autoref{tab:examples}, which contains \colorbox{neu}{\textit{Neutral}} examples where the hypothesis introduces an exclusivity that is not implied by the premise.}
    \label{tab:row2}
\end{table*}
\begin{table*}[]
    \centering
    \resizebox{\textwidth}{!}{%
    \begin{tabular}{l}
    \toprule
    \makecell[l]{
    Write a pair of sentences that have the same relationship as the previous examples. Examples:\\[4pt]
    1. Dun Laoghaire is the major port on the \textit{\textbf{south coast}}.\\
    Contradiction: Dun Laoghaire is the major port on the \textit{\textbf{north coast}}.\\[4pt]
    2. Leave the city by its \textit{\textbf{eastern Nikanor Gate}} for a five-minute walk to Hof Argaman (Purple Beach), one\\of Israel's finest beaches.\\
    Contradiction: Leave the city by its \textbf{\textit{western Nikanor Gate}} for a fifty five minute walk to Hof Argaman.\\[4pt]
    3. \textit{\textbf{Southwest of the Invalides}} is the Ecole Militaire, where officers have trained since the middle of the \\
    18th century.\\
    Contradiction: \textit{\textbf{North of the Invalides}} is the Ecole Militaire, where officers have slept since the early 16th\\ century. \\[4pt]
    4. Across the courtyard on the \textit{\textbf{right-hand side}} is the chateau's most distinctive feature, the splendid \\
    Francois I wing.\\
    Contradiction: The Francois l wing can be seen across the courtyard on the \textit{\textbf{left-hand side}}.\\[4pt]
    5. \textit{\textbf{To the south}}, in the Sea of Marmara, lie the woods and beaches of the Princes' Islands.\\
    Contradiction: \textit{\textbf{In the north}} is the Sea of Marmara where there are mountains to climb.\\[4pt]
    6.}\\
    \bottomrule
    \end{tabular}}
    \caption{Context corresponding to row 3 in \autoref{tab:examples}, which contains \colorbox{con}{\textit{Contradiction}} examples that flip cardinal directions between the premise and hypothesis.}
    \label{tab:row3}
\end{table*}
\begin{table*}[]
    \centering
    \resizebox{\textwidth}{!}{%
    \begin{tabular}{l}
    \toprule
    \makecell[l]{
    Write a pair of sentences that have the same relationship as the previous examples. Examples:\\[4pt]
    1. Vendors and hair braiders are sure to \textit{\textbf{approach}} you.\\
    Implication: You're likely to be \textit{\textbf{solicited by}} vendors or hair braiders.\\[4pt]
    2. The Carre d'Art, an ultramodern building opposite the Maison Carre, \textit{\textbf{exhibits}} modern art.\\
    Implication: Pieces of modern art \textit{\textbf{can be found}} in the Carre d'Art, a structure which stands \\
    across from the Maison Carre. \\[4pt]
    3. But they also take pains not to dismiss the trauma the Holocaust visited and continues to visit upon Jews.\\
    Implication: The Holocaust visited trauma upon Jews, and they are careful not to dismiss this. \\[4pt]
    4. One fortunate \textit{\textbf{result}} of this community's influence has been the proliferation of good restaurants \\
    and interesting bars from which to choose.\\
    Implication: The influence of this community has \textit{\textbf{led to}} an increase in the number of intriguing bars and \\
    good dining establishments.\\[4pt]
    5. Salinger \textit{\textbf{wrote}} similar letters \textit{\textbf{to}} other young female writers.\\
    Implication: Other young female writers \textit{\textbf{received}} similar letters \textit{\textbf{from}} Salinger as well.\\[4pt]
    6.
    }\\
    \bottomrule
    \end{tabular}}
    \caption{Context corresponding to row 7 in \autoref{tab:examples}, which contains \colorbox{ent}{\textit{Entailment}} examples that substitute a verb in the premise with one in the hypothesis that has a different subcategorization frame. Note that the third in-context example does not share quite the same pattern, but GPT-3 is still able to replicate the pattern present in other examples.}
    \label{tab:row7}
\end{table*}

\end{document}